%% file: paper.tex
\documentclass[10pt,journal,compsoc]{IEEEtran}

\ifCLASSOPTIONcompsoc
  \usepackage[nocompress]{cite}
\else
  \usepackage{cite}
\fi

\usepackage{subfigure}

\usepackage{amsthm}
\usepackage{amsmath,amsfonts}
\usepackage{amssymb}

\usepackage{array}
\usepackage{textcomp}
\usepackage{stfloats}
\usepackage{url}
\usepackage{verbatim}
\usepackage{graphicx}

\usepackage[ruled,linesnumbered,vlined]{algorithm2e}
\SetKwInOut{Input}{Input}
\SetKwInOut{Output}{Output}
\usepackage{algorithmic}

\usepackage{multirow}
\usepackage{booktabs}
\usepackage{colortbl}
\usepackage{enumitem}
\usepackage{xspace}
\usepackage{xcolor}
\usepackage{makecell}

\usepackage[capitalize]{cleveref}

\newcommand{\inlinegraphics}[1]{\raisebox{-.1\height}{\includegraphics[width=0.95em]{#1}}}

\begin{document}

\title{Bi-Level Graph Structure Learning for\\ Next POI Recommendation}

\author{Liang~Wang,
        Shu~Wu,~\IEEEmembership{Senior Member,~IEEE,}
        Qiang~Liu,~\IEEEmembership{Member,~IEEE,}
        Yanqiao~Zhu,
        Xiang~Tao,
        Mengdi~Zhang,
        and~Liang~Wang,~\IEEEmembership{Fellow,~IEEE,}
\IEEEcompsocitemizethanks{\IEEEcompsocthanksitem L. Wang, S. Wu, Q. Liu, X. Tao, and L. Wang are with the Center for Research on Intelligent Perception and Computing (CRIPAC), State Key Laboratory of Multimodal Artificial Intelligence Systems (MAIS), Institute of Automation, Chinese Academy of Sciences (CASIA), and also with the School of Artificial Intelligence, University of Chinese Academy of Sciences, Beijing 100190, China. Email: \{liang.wang, xiang.tao\}@cripac.ia.ac.cn, \{shu.wu, qiang.liu, wangliang\}@nlpr.ia.ac.cn.
\IEEEcompsocthanksitem Y. Zhu is with the Department of Computer Science, University of California, Los Angeles, Los Angeles, CA 90095 USA.\protect\\
Email: yzhu@cs.ucla.edu.
\IEEEcompsocthanksitem M. Zhang is with the Meituan, Beijing 100102, China.\protect\\
Email: zhangmengdi02@meituan.com
\IEEEcompsocthanksitem Corresponding author: Shu Wu.}
\thanks{Manuscript received 22 June 2023; revised 23 April 2024; accepted 29 April 2024.}}

\markboth{Journal of \LaTeX\ Class Files,~Vol.~14, No.~8, August~2015}%
{Shell \MakeLowercase{\textit{et al.}}: Bare Demo of IEEEtran.cls for Computer Society Journals}
%


\newcommand{\themodel}{\textsf{BiGSL}\xspace}

\IEEEtitleabstractindextext{
\begin{abstract}
Next point-of-interest (POI) recommendation aims to predict a user’s next destination based on sequential check-in history and a set of POI candidates. Graph neural networks (GNNs) have demonstrated a remarkable capability in this endeavor by exploiting the extensive global collaborative signals present among POIs. However, most of the existing graph-based approaches construct graph structures based on pre-defined heuristics, failing to consider inherent hierarchical structures of POI features such as geographical locations and visiting peaks, or suffering from noisy and incomplete structures in graphs. To address the aforementioned issues, this paper presents a novel \underline{Bi}-level \underline{G}raph \underline{S}tructure \underline{L}earning (\themodel) for next POI recommendation. \themodel first learns a hierarchical graph structure to capture the fine-to-coarse connectivity between POIs and prototypes, and then uses a pairwise learning module to dynamically infer relationships between POI pairs and prototype pairs. Based on the learned bi-level graphs, our model then employs a multi-relational graph network that considers both POI- and prototype-level neighbors, resulting in improved POI representations. Our bi-level structure learning scheme is more robust to data noise and incompleteness, and improves the exploration ability for recommendation by alleviating sparsity issues. Experimental results on three real-world datasets demonstrate the superiority of our model over existing state-of-the-art methods, with a significant improvement in recommendation accuracy and exploration performance.
\end{abstract}

}

\maketitle

\IEEEdisplaynontitleabstractindextext

%
\IEEEpeerreviewmaketitle

\input{sections/introduction.tex}
\input{sections/related-work.tex}
\input{sections/preliminaries.tex}
\input{sections/method.tex}
\input{sections/experiments.tex}
\input{sections/conclusion.tex}

\ifCLASSOPTIONcompsoc
  \section*{Acknowledgments}
\else
  \section*{Acknowledgment}
\fi

This work was supported by National Natural Science Foundation of China (62206291, 62141608).

\ifCLASSOPTIONcaptionsoff
  \newpage
\fi

\bibliographystyle{IEEEtran}
\bibliography{paper}

\vspace{-10 pt}
\begin{IEEEbiography}[{\includegraphics[width=1in,height=1.25in,clip,keepaspectratio]{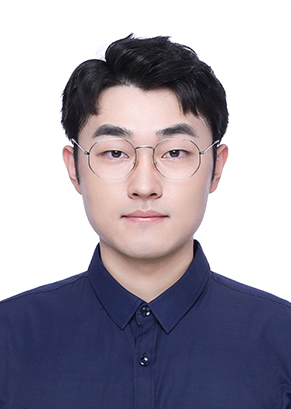}}]{Liang Wang}
is currently pursuing his Ph.D. degree of Computer Science at the Center for Research on Intelligent Perception and Computing (CRIPAC) at State Key Laboratory of Multimodal Artificial Intelligence Systems (MAIS), Institute of Automation, Chinese Academy of Sciences (CASIA). His current research interests mainly include graph representation learning, AI for science, and spatio-temporal data mining.
\end{IEEEbiography}
\vspace{-10 pt}

\begin{IEEEbiography}[{\includegraphics[width=1in,height=1.25in,clip,keepaspectratio]{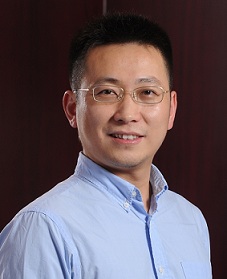}}]{Shu Wu}
received his B.S. degree from Hunan University, China, in 2004, M.S. degree from Xiamen University, China, in 2007, and Ph.D. degree from Department of Computer Science, University of Sherbrooke, Quebec, Canada, all in computer science. He is an Associate Professor with the Center for Research on Intelligent Perception and Computing (CRIPAC), State Key Laboratory of Multimodal Artificial Intelligence Systems (MAIS), Institute of Automation, Chinese Academy of Sciences (CASIA).
He has published more than 50 papers in the areas of data mining and information retrieval in international journals and conferences, such as IEEE TKDE, IEEE THMS, AAAI, ICDM, SIGIR, and CIKM. His research interests include data mining, information retrieval, and recommendation.
\end{IEEEbiography}
\vspace{-10 pt}

\begin{IEEEbiography}[{\includegraphics[width=1in,height=1.25in,clip,keepaspectratio]{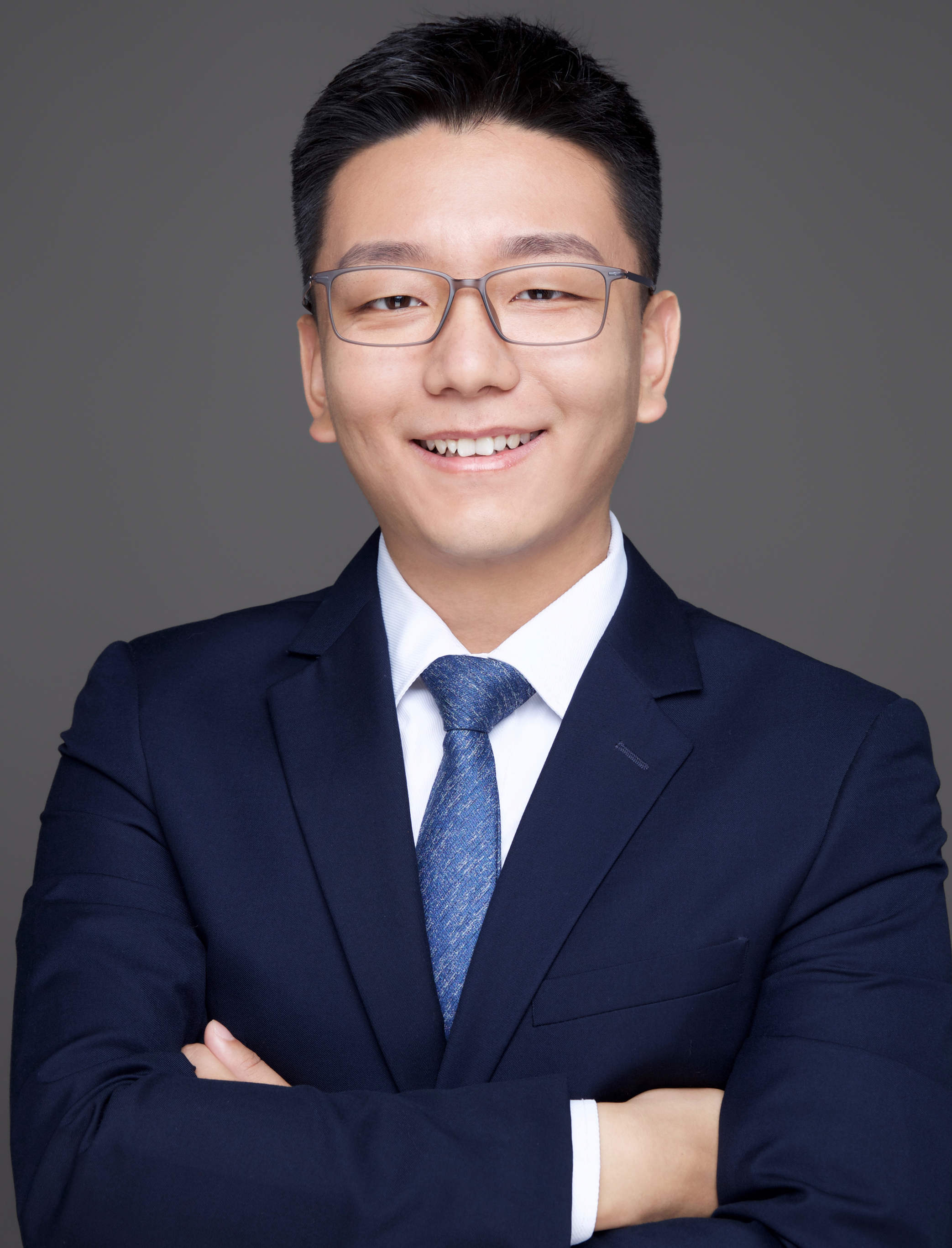}}]{Qiang Liu}
is an Associate Professor with the Center for Research on Intelligent Perception and Computing (CRIPAC), State Key Laboratory of Multimodal Artificial Intelligence Systems (MAIS), Institute of Automation, Chinese Academy of Sciences (CASIA). He received his PhD degree from CASIA. Currently, his research interests include data mining, misinformation detection, LLM safety and AI for science. He has published papers in top-tier journals and conferences, such as IEEE TKDE, AAAI, NeurIPS, KDD, WWW, SIGIR, CIKM, ICDM, ACL and EMNLP.
\end{IEEEbiography}
\vspace{-10 pt}

\begin{IEEEbiography}[{\includegraphics[width=1in,height=1.25in,clip,keepaspectratio]{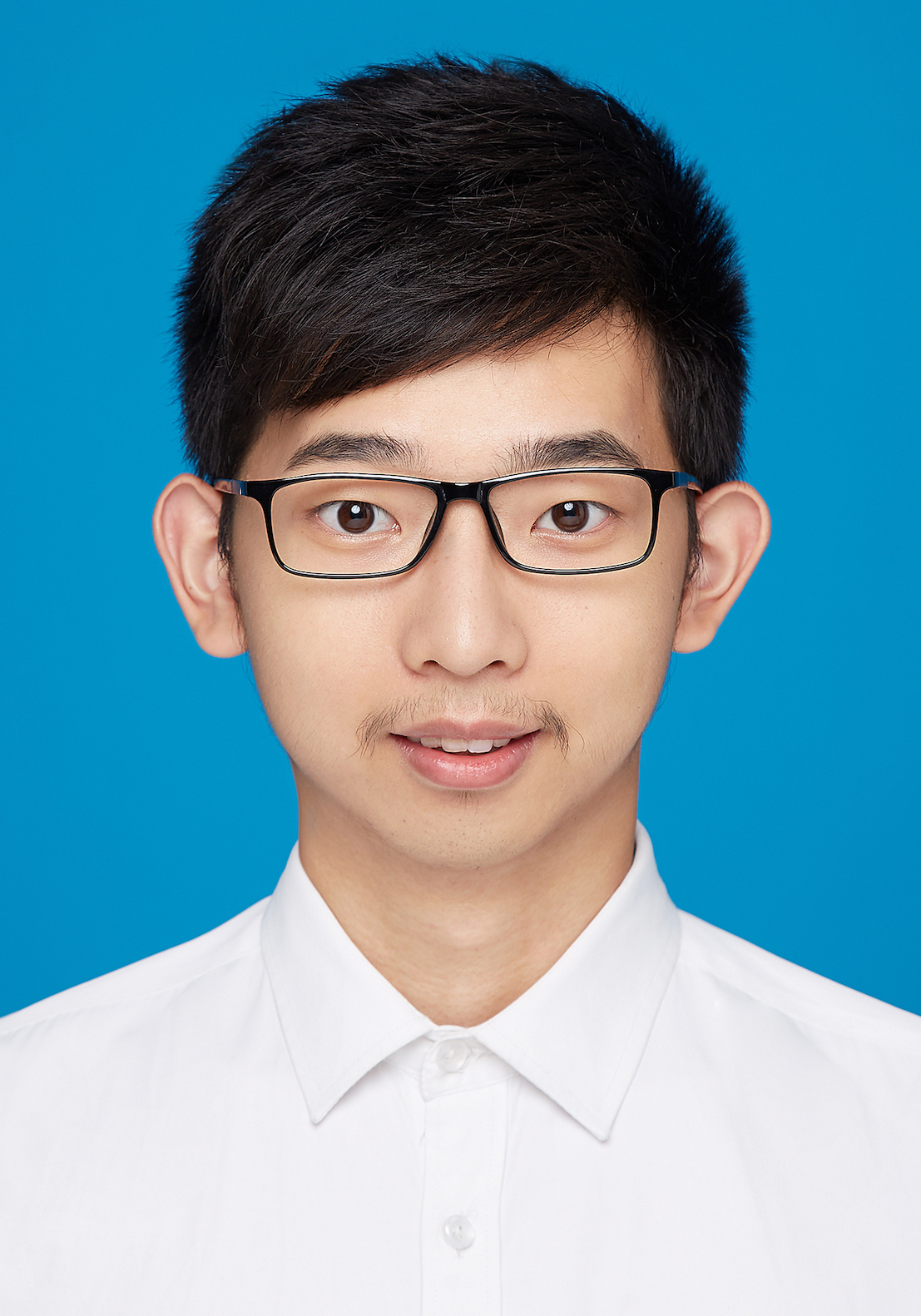}}]{Yanqiao Zhu}
is currently pursuing his Ph.D. degree at the Department of Computer Science, University of California, Los Angeles. His current research interests involve the exploration of graph and geometric representation learning and particularly their applications in computational chemistry.
\end{IEEEbiography}
\vspace{-10 pt}

\begin{IEEEbiography}[{\includegraphics[width=1in,height=1.25in,clip,keepaspectratio]{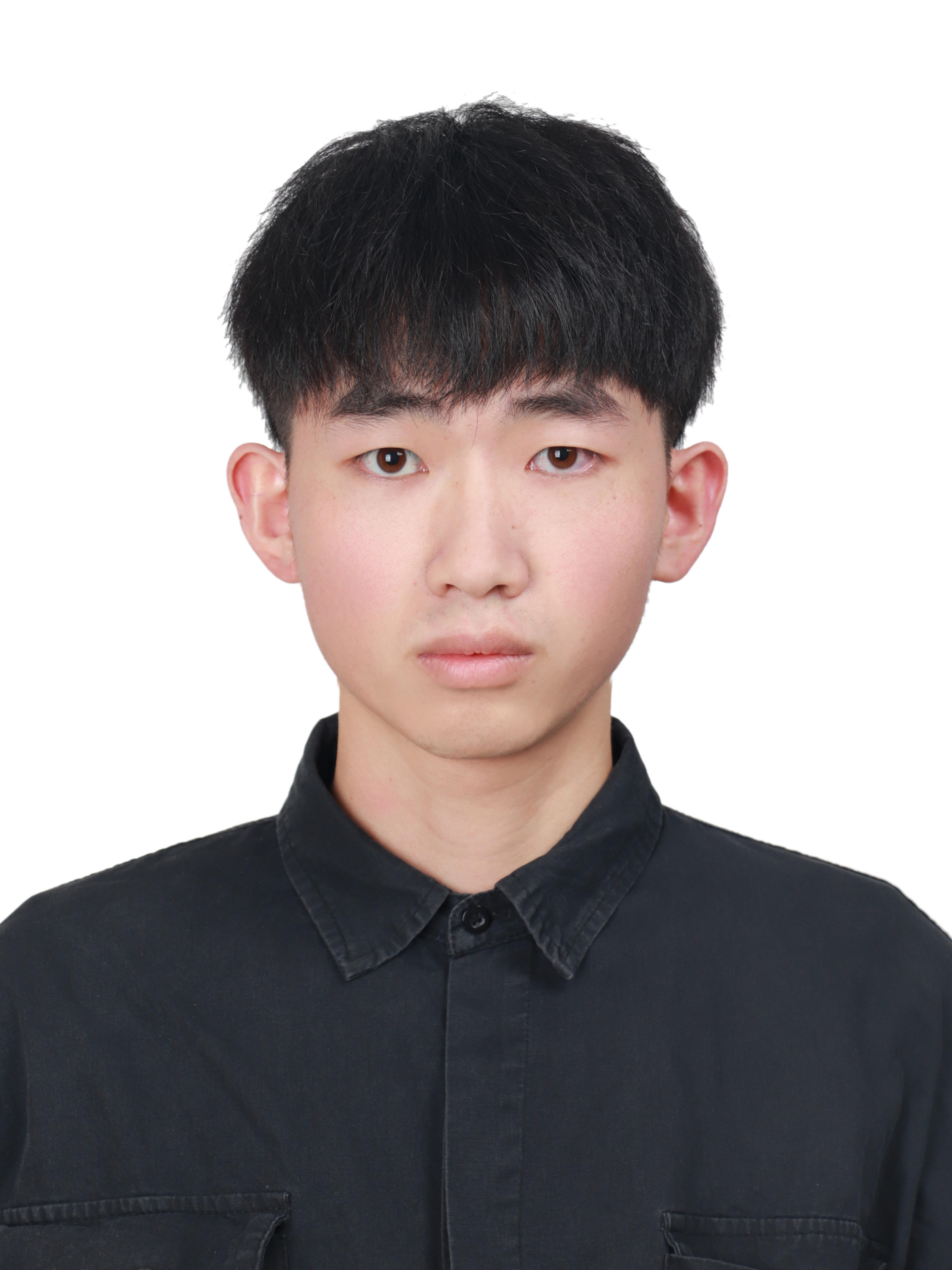}}]{Xiang Tao}
is currently pursuing his master’s degree of Computer Science at the Center for Research on Intelligent Perception and Computing (CRIPAC) at State Key Laboratory of Multimodal Artificial Intelligence Systems (MAIS), Institute of Automation, Chinese Academy of Sciences (CASIA). His current research interests mainly include data mining, graph representation learning and recommender systems.
\end{IEEEbiography}
\vspace{-10 pt}

\begin{IEEEbiography}[{\includegraphics[width=1in,height=1.25in,clip,keepaspectratio]{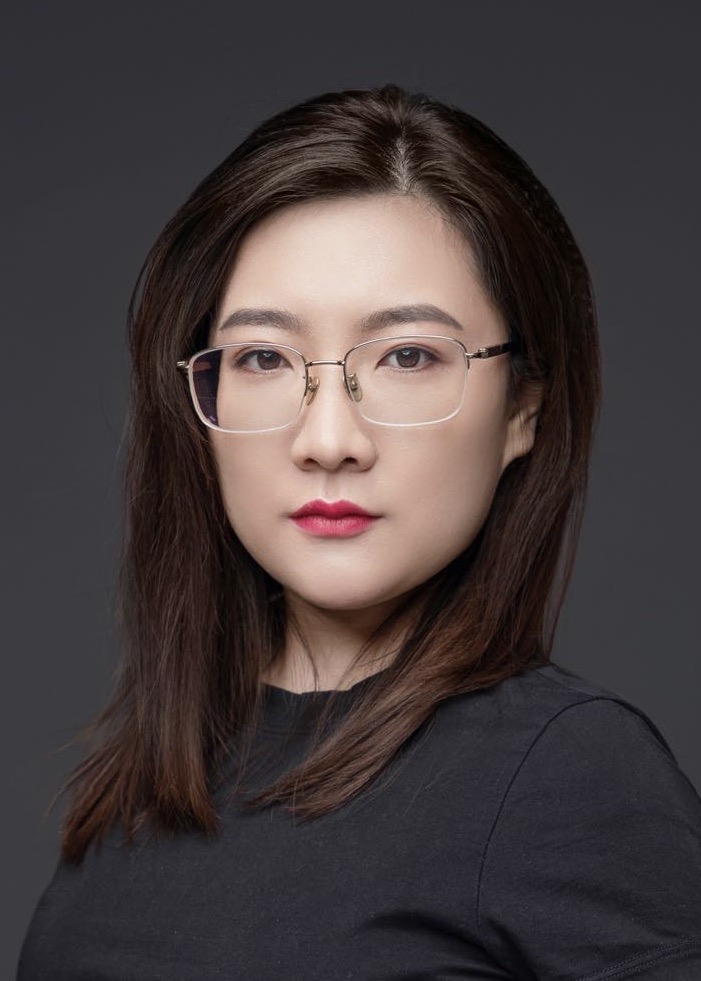}}]{Mengdi Zhang}
is a senior algorithm expert and technical manager at Meituan. She co-founded the OpenKG community and serves as a committee member of CIPS SIGKG. She received the BEng degree from Shandong University in 2014. Her research interests lie in industry large-scale knowledge representation and reasoning systems. She has published papers in top journals and conferences, such as IEEE TKED, AAAI, IJCAI, ACL, SIGIR, KDD, and WWW. She was a winner of the 12th Wu Wenjun AI Science and Technology Progress Award.
\end{IEEEbiography}
\vspace{-10 pt}

\begin{IEEEbiography}[{\includegraphics[width=1in,height=1.25in,clip,keepaspectratio]{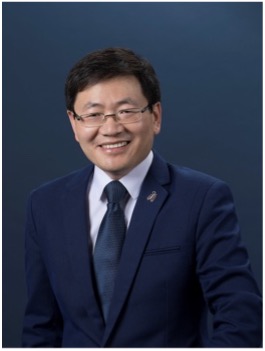}}]{Liang Wang}
received both the BEng and MEng degrees from Anhui University in 1997 and 2000, respectively, and the PhD degree from the Institute of Automation, Chinese Academy of Sci- ences (CASIA) in 2004. From 2004 to 2010, he was a research assistant at Imperial College London, United Kingdom, and Monash University, Australia, a research fellow at the University of Melbourne, Australia, and a lecturer at the University of Bath, United Kingdom, respectively. Currently, he is a full professor of the Hundred Talents Program at the State Key Laboratory of Multimodal Artificial Intelligence Systems, CASIA. His major research interests include machine learning, pattern recognition, and computer vision. He has widely published in highly ranked international journals such as IEEE TPAMI and IEEE TIP, and leading international conferences such as CVPR, ICCV, and ECCV. He has served as an Associate Editor of IEEE TPAMI, IEEE TIP, and PR. He is an IEEE Fellow and an IAPR Fellow.
\end{IEEEbiography}

\end{document}

%% file: sections/introduction.tex
\IEEEraisesectionheading{\section{Introduction}}

\IEEEPARstart{T}{he} emergence of location-based social networks has brought to light a subject of great interest to both researchers and service providers alike: next point-of-interest (POI) recommendation.
This task seeks to comprehend the temporal nature of a user's preferences by analyzing their historical check-in sequences and then make predictions about the next POIs that they are most likely to visit. Such insights can be used to improve both the user experience as well as the service provider's services.

Graph-based methods have been widely used in POI recommendation due to their capability of modeling global collaborative relationships of POIs across users. These methods typically involve two stages: (1) the construction of a topology graph based on POI features and (2) the learning of POI representations based on the constructed graph.
Depending on the type of information to be used, such a graph may be built by taking into consideration the spatial information of POIs, such as distance intervals~\cite{STP-UDGAT,Graph-Flashback,DRAN} or grid regions~\cite{HMT-GRN}, as well as temporal features from users' sequential check-in data, such as the average time intervals between consecutive visits~\cite{STP-UDGAT} or the Jaccard similarity of time slot sets~\cite{HMT-GRN}.
Additionally, it is also common to model transitions between POIs based on the number or frequency of consecutive visits between each POI~\cite{SGRec, GETNext, DRAN}.
After the graph is built, graph neural networks (GNNs) are used to learn the POI representations by aggregating information from the neighborhood of the nodes. These POI representations are then used to further learn users’ preferences from the sequences of visited POIs and rank candidate POIs for producing recommendations.

Despite their success, existing graph-based methods for POI recommendation suffer from various limitations.

\begin{figure}
    \centering
    \includegraphics[width=\linewidth]{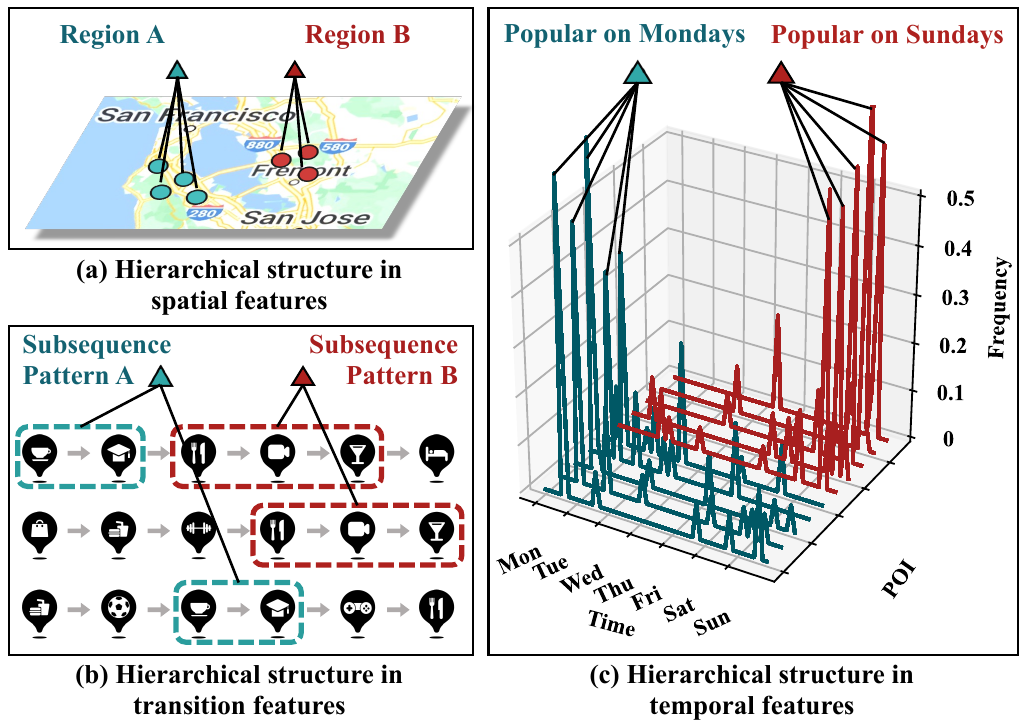}
    \caption{Hierarchical structures in different POI features, i.e., \textit{fine-grained POIs} can be divided into \textit{coarse-grained groups} (corresponding to cyan and red triangles in the figure). We use prototypes to represent coarse-grained group information and introduce prototype nodes in the graph structure learning to construct hierarchical graphs.}
    \label{fig:hierarchical_structure}
\end{figure}

\textbf{\textit{Firstly}}, previous methods construct graphs based solely on local neighborhoods, disregarding the valuable hierarchical structures of POIs.
Hierarchical structure means that fine-grained POIs can be divided into coarse-grained groups, and POIs within the same group have similar group characteristics in some aspects.
As shown in \cref{fig:hierarchical_structure}, 
POIs could be grouped into the same group due to similar spatial locations, transition sequence patterns, temporal visiting peaks, or category descriptors.
These hierarchical structures have been proven to improve recommendations by mitigating the sparsity issues~\cite{ASPPA,GeoSAN,HGMF} and improving the exploration ability~\cite{HMT-GRN,HME}.
Some previous methods handle hierarchical information by employing multi-task learning~\cite{HMT-GRN} or designing hierarchical encoders~\cite{STAR}, however, the ingenious combination of the advantages of GNNs and hierarchical structures remains unexplored.

\textbf{\textit{Secondly}}, the graph structure of these existing methods is usually fixed during training, determined by pre-defined rules or empirical laws~\cite{GETNext, DRAN, ARNN, STGCAN}. This lack of flexibility could lead to performance degradation due to the presence of noise or incompleteness in the graph structure, since GNNs heavily rely on the quality of graph structure~\cite{gsl-survey}.
For example, there could be noisy and missing edges in the graph structure, due to erroneous check-ins and missed check-ins by users, or even malicious attacks.

\textbf{\textit{Lastly}}, multiple POI features (e.g., spatial, transition, and temporal features) are usually leveraged to construct multiple graphs.
In different graphs, the same POI may have common neighbors and graph-specific neighbors, so there is shared information and specific information contained in different POI features.
However, existing approaches often simply linearly combine or concatenate the POI representations in each graph to fuse them~\cite{STP-UDGAT,HMT-GRN}, which inadequately models the alignment and complementary relationships of POI representations in different graphs, and results in sub-optimal POI representations.

To overcome the aforementioned limitations, we introduce a novel \underline{Bi}-level \underline{G}raph \underline{S}tructure \underline{L}earning method for next POI recommendation, \themodel for brevity.
Our proposed \themodel model consists of two graph structure learning modules that adaptively capture coarse- and fine-grained connectivity structures of POIs.
Specifically, we first map each POI to a node in the graph space and then resort to clustering on the POI features to discover the hierarchical structure. The resulting prototypes represent the coarse-grained clusters, which are then added to the graph to augment the neighborhood of POI nodes.
Subsequently, we introduce a pairwise structure learning method to infer the connectivity between POI pairs and prototype pairs in an adaptive manner, the result of which represents fine-grained connections that supplement the coarse-grained information.
This bi-level approach produces coarse-to-fine connectivities of POIs that can be learned in a data-driven manner, thereby alleviating the first two limitations.
Based on the bi-level graphs, we propose a multi-relational graph attention network that considers two facets of local structures including POI- and prototype-level neighbors, which produces better POI representations as a result.
Finally, to further boost the performance, we construct multiple views based on the original POIs and users' sequential data. To encourage the fusion of POI information from distinct views, we design a contrastive multiview fusion approach by mining view-shared and view-specific information, which better aligns complementary features in different views.

The main contributions of our work are outlined as follows:
\begin{itemize}
    \item We propose a novel \themodel model for next POI recommendation, which employs a bi-level graph structure learning method that adaptively infers hierarchical graph structures in a data-driven manner.
    \item Based on the learned bi-level graphs, we design a multi-relational graph network to generate more informative POI representations, considering both POI- and prototype-level neighbors.
    \item We further introduce a multiview contrastive learning strategy to integrate information from multiple views to improve the recommendation accuracy.
    \item We conduct extensive experiments on three real-world datasets including two widely adopted benchmarks (Gowalla and Foursquare) and a new commercial dataset. The results show that \themodel significantly outperforms state-of-the-art methods in recommendation accuracy and exploration performance.
\end{itemize}

%% file: sections/related-work.tex
\section{Related Work}

In this section, we succinctly review existing studies for next POI recommendation and graph structure learning.

\subsection{Next POI Recommendation}

Next POI recommendation aims to infer the users' dynamic preferences and predict where the user will go next, given the historical check-ins and a set of POI candidates. 

Recurrent neural networks (RNNs) and self-attention have shown promising performance in handling sequential data, hence they have been widely used as the backbone of the next POI recommenders~\cite{SNPR,TLR-M,MFNP,CHAML}.
Some studies are dedicated to capturing sequential dependencies in sequences to model the dynamic user preferences~\cite{LSTPM,PLSPL,CFPRec,TA-RLBL}.
On the other hand, POI features and historical check-ins contain rich collaborative signals, such as spatial location, visited time, and frequency of consecutive visits, and are therefore leveraged to make more effective recommendations from sparse data.
Early works introduce these collaborative signals directly into the backbone, such as computing transition matrices or gates in RNNs~\cite{ST-RNN,STGN,NeuNext,Flashback} and attention maps in self-attention~\cite{STAN}.

Recently, it has been found that these collaborative signals can be represented by graphs and the GNNs can be employed to effectively capture the correlations among POIs~\cite{ARNN,DRAN,STGCAN}.
For example, STP-UDGAT~\cite{STP-UDGAT} constructs three types of POI graphs based on the spatial distance, time interval, and consecutive visiting, so as to learn user preferences in different views. 
GETNext~\cite{GETNext} introduces a global trajectory graph to better leverage the extensive collaborative signals from different users. 
HMT-GRN~\cite{HMT-GRN} constructs global spatio-temporal graphs to model collaborative signals among POIs and utilizes auxiliary tasks to alleviate the data sparsity issue.

Although great success has been achieved, these existing graph-based methods rely on pre-defined rules to construct graphs, which leads to noise and incompleteness in the graph and degenerates the performance of GNNs~\cite{iMTL}. Although GraphFlashback~\cite{Graph-Flashback} attempts to automatically learn latent POI graphs based on a holistic knowledge graph, the topology of such latent graphs lacks interpretability and still neglects the underlying hierarchical structure in POI features.
SNPM~\cite{SNPM} is another attempt to infer latent graph structures. However, it only considers relationships within the same region and remains constrained by a pre-defined graph structure.
Our proposed \themodel addresses the above issues by hierarchical and pairwise structure learning.

\subsection{Graph Structure Learning}

Although GNNs have achieved superior performance in analyzing graph-structured data, most GNNs are highly sensitive to the quality of graph structures and usually require a credible graph structure that is hard to construct in real-world applications~\cite{gsl-discrete}. Given that GNNs recursively aggregate information from neighborhoods to update node embeddings, the iterative nature of this process has consequential cascading effects. Small noise in a graph will be propagated to the neighboring nodes, subsequently affecting the embeddings of numerous other nodes~\cite{gsl-survey,GSLB}. Recently, considerable literature has arisen around the theme of graph structure learning (GSL), which targets at jointly learning an optimized graph structure and corresponding representations. 
Existing GSL methods can be roughly grouped into three categories: metric learning that models or refines the edge weights by measuring the similarity between node representations~\cite{gsl-ml1,gsl-ml2,gsl-ml3}, probabilistic modeling that models the probability distribution of edges and then samples a graph from this distribution~\cite{gsl-discrete,gsl-pm2,gsl-pm3}, and direct optimization that treats the graph adjacency matrix as parameters and optimizes it directly along with GNN parameters~\cite{gsl-do1,gsl-do2,gsl-do3}.

Although these studies leverage graph structure learning to refine the graph structures, they are not tailored for next POI recommendation. They only learn pairwise relationships between nodes and lack consideration on meaningful hierarchical structure. NCL~\cite{NCL} extracts coarse-grained prototypes for cross-granularity contrastive learning in graph collaborative filtering, inspiring us to construct hierarchical graph structures using prototypes.
Distinct from previous graph structure learning methods, we not only adaptively learn pairwise relationships between POIs to suppress the potential noise, but also construct hierarchical structures to further enrich the neighborhoods of POIs by extracting POI prototypes, so as to capture the global relationships between POIs effectively and provide accurate recommendations.

%% file: sections/preliminaries.tex
\section{Preliminaries and Problem Statement}

\begin{figure*}[htb]
\centering
\includegraphics[width=\linewidth]{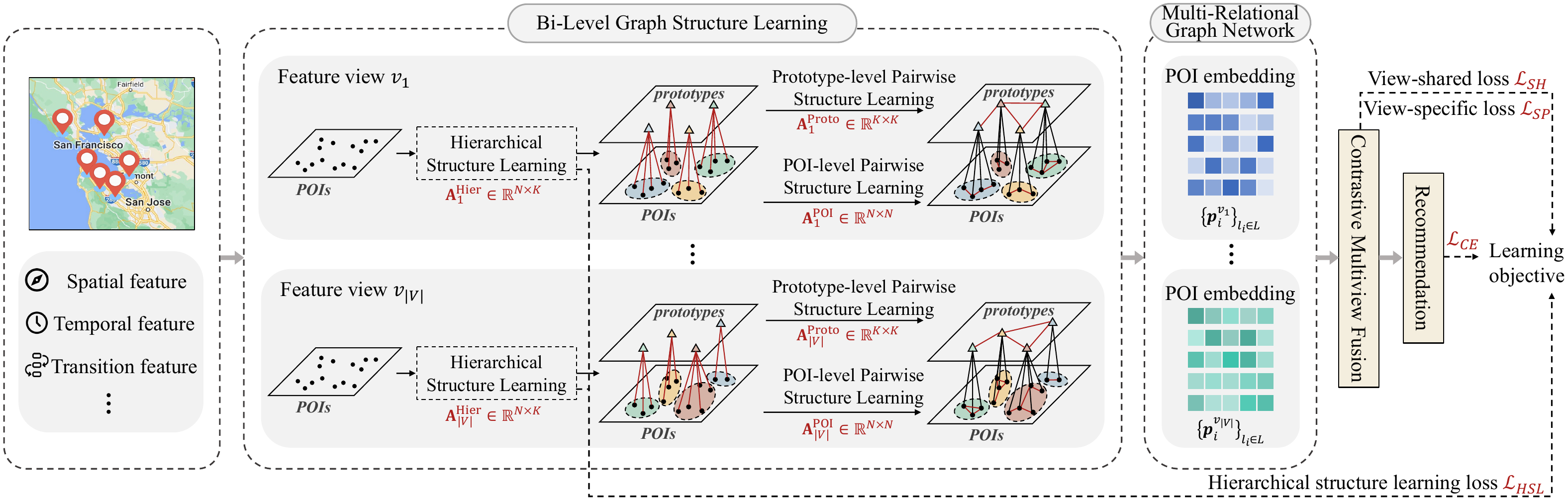}
\caption{The overall framework of our proposed \themodel model. We first construct multiple feature views from primitive POI features. In each view, we map POIs to nodes and then cluster the POI features to reveal the hierarchical structure. The resulting prototypes represent the coarse-grained group information, which are added to the graph.
Subsequently, we perform pairwise structure learning to infer the connectivity between POI pairs and prototype pairs, resulting in a data-driven hierarchical graph.
Based on the hierarchical graph, we conduct the multi-relational graph learning to produce better POI representations.
Finally, to encourage POI information fusion from different views and make better recommendations, we adopt a contrastive multiview fusion approach by mining view-shared and view-specific information.}
\label{overview}
\end{figure*}

Let $U=\{u_1, u_2, \dots,u_M\}$ and $L = \{l_1, l_2, \dots, l_N\}$ denote the set of $M$ users and $N$ POIs, respectively.

\textbf{Definition 1. Check-in.} 
A check-in record is represented in a tuple $(l_{t_i}, loc_{t_i}, time_{t_i})$, in which $l_{t_i}$ is the POI visited on time step $t_i$ with its location coordinates $loc_{t_i}$, and $time_{t_i}$ is the timestamp. 

\textbf{Definition 2. Check-in Sequence.}
Each user $u_m$ has a chronologically ordered historical check-in sequence $s_{m} = \{(l_{t_1}, loc_{t_1}, time_{t_1}), (l_{t_2}, loc_{t_2}, time_{t_2}), \dots, (l_{t_{i}}, loc_{t_{i}}, time_{t_{i}})\}$, where $l_{t_i}$ is the last POI visited.

\textbf{Definition 3. POI Feature.}
Apart from the POI IDs, POIs typically possess several primitive features, including spatial features (latitude and longitude), temporal features (the time distribution of check-ins), transition features (the distribution of consecutive check-ins between POIs), and category features, among others. These features reveal the similarity or collaborative relationships among POIs, making them useful in constructing graphs for existing graph-based next POI recommendation methods, following pre-defined rules. For instance, a spatial graph can be constructed by dividing grids according to latitude and longitude or calculating distances, while a temporal graph can be constructed based on temporal feature similarity.

\textbf{Definition 4. Feature View.}
To effectively capture collaborative signals embedded in different primitive features, we adopt a multi-view framework. In each feature view, we model the information provided by only one type of primitive features and then fuse the information from all views. We denote the set of feature views, which also represents the set of POI primitive feature types, as $V$.
In feature view $v\in V$, the primitive features of POIs are denoted by $X^{v}\in\mathbb{R}^{N\times d_1^v}$, where $d_1^v$ represents the dimension of the primitive feature. The $i$-th row, $x_i\in\mathbb{R}^{d_1^v}$, denotes the primitive feature of POI $l_i$.
For instance, in the spatial feature view, $\boldsymbol{x}_i^{\mathrm{spatial}} = (latitude_i, longitude_i)$, and $d_1^{\mathrm{spatial}} = 2$.

\textbf{Problem 1. Next POI Recommendation.}
Next POI recommendation takes a user's historical check-in sequence $s_{m}$ and the POI candidate set $L$ as input to generate a ranked POI list for the next time step $t_{i+1}$, where the next visited POI $l_{t_{i+1}}$ should be highly ranked.

%% file: sections/method.tex
\section{The proposed \themodel Model}

This section elaborates on our proposed \themodel. We first introduce the base recommender backbone.
Then, we introduce four main components in our model: (1) hierarchical structure learning that infers the hierarchical structure by grouping POIs into different clusters and extracting prototypes, (2) pairwise structure learning that adaptively infers relationships between POIs or prototypes, (3) a novel multi-relational graph attention network that can fully exploit learned hierarchical graphs, and (4) contrastive multiview fusion that computes view-shared and view-specific representations to facilitate information fusion.
Finally, we explain how to optimize our model and provide a complexity analysis.
The overall framework of \themodel is shown in \cref{overview}.

\subsection{Backbone Recommender}\label{backbone}

The key components of our proposed \themodel are model-agnostic and can be plugged into any sequential recommendation model.
To show the effectiveness of our approach, we choose a simple yet effective recommender as the backbone, which contains only an embedding layer, an LSTM layer and a dense layer with softmax normalization.

The embedding layer offers dense POI ID embeddings and user ID embeddings. We describe a POI $l_i$ (a user $u_m$) with an embedding vector $\boldsymbol{l}_i \in \mathbb{R}^{d_2}$ ($\boldsymbol{u}_m \in \mathbb{R}^{d_2}$), where $d_2$ denotes the embedding size.

We first use the LSTM layer to learn the user's dynamic preference from the historical check-in sequence $s_{m}$:
\begin{equation}
\boldsymbol{h}_{t_{i}}=\operatorname{LSTM}\left(s_{m}\right),
\end{equation}
\noindent where $\operatorname{LSTM}(\cdot)$ represents the LSTM layer,
and $\boldsymbol{h}_{t_{i}} \in \mathbb{R}^{d_3}$ is the hidden representation of user $u_m$'s historical check-ins.

Next, we compute the conditional probability of next POI distribution and rank all POIs to make personalized recommendation:
\begin{equation}
    \hat{\boldsymbol{y}} = \operatorname{Softmax}\left(\mathbf{W}\left(\boldsymbol{h}_{t_{i}} || \boldsymbol{u}_m\right)\right),
\end{equation}
\noindent where $\hat{\boldsymbol{y}} \in \mathbb{R}^N$ is the predicted conditional probability distribution regarding $t_{i+1}$, and $\hat{\boldsymbol{y}}_i = P(l_{i}|s_{m})$ is the probability that the next POI is $l_i$ given the historical sequence $s_m$. $\boldsymbol{u}_m \in \mathbb{R}^{d_2}$ is the trainable user ID embedding to introduce personalization, $||$ is the vector concatenation, and $\mathbf{W}\in \mathbb{R}^{N \times (d_2+d_3)}$ is the learnable weight matrix. The softmax function is performed to compute the conditional probability of the next POI distribution. Finally, we can sort POIs in descending order of conditional probability and get a ranked POI list.

\subsection{Hierarchical Structure Learning}\label{sec:hierarchical}

Based on the aforementioned backbone, we point out that existing methods, which employ conventional GNNs on manually constructed graphs to encode the interaction relationship between POIs, cannot model the hierarchical information well. 
To address this concern, we propose a hierarchical structure learning method to embed the hierarchical information in the graph structures explicitly.

To model the hierarchical nature in POI features, we propose a hierarchical structure learning method. We design a hierarchical structure learning objective to group POIs into different clusters and extract prototypes. Roughly speaking, prototypes can be regarded as the center of clusters that represent a group of semantically similar POIs.

Formally, the goal of graph structure learning is to maximize the following log-likelihood function:
\begin{equation}
\label{prototype}
\begin{aligned}
    L &= \sum_{l_i \in L} \log p\left(\boldsymbol{z}_i \mid \mathbf{\Theta}, \mathbf{X}\right) \\
    &=\sum_{l_i \in L} \log \sum_{\boldsymbol{c}_j \in C} p\left(\boldsymbol{z}_i, \boldsymbol{c}_j \mid \mathbf{\Theta}, \mathbf{X}\right),
\end{aligned}
\end{equation}
where $\mathbf{\Theta}$ is learnable parameters of model, $\mathbf{X}$ is primitive features of POIs, $\boldsymbol{z}_i \in \mathbb{R}^{d_2}$ is the learned structure embedding of POI $l_i$ that will be used to construct the graph structure. $C$ is the set of cluster centroids, and we use $K = |C|$ to denote the number of clusters.
The objective in \cref{prototype} is hard to optimize directly because $\boldsymbol{z}_i, \boldsymbol{c}_j$ are both free variables. Therefore, we introduce its tractable lower bound by Jensen’s inequality:
\begin{equation}
\begin{aligned}
L=&\sum_{l_i \in L} \log \sum_{\boldsymbol{c}_j \in C} Q\left(\boldsymbol{c}_j \mid \boldsymbol{z}_i\right) \frac{p\left(\boldsymbol{z}_i, \boldsymbol{c}_j \mid \mathbf{\Theta}, \mathbf{X}\right)}{Q\left(\boldsymbol{c}_j \mid \boldsymbol{z}_i\right)} \\
 \geq &\sum_{l_i \in L} \sum_{\boldsymbol{c}_j \in C} Q\left(\boldsymbol{c}_j \mid \boldsymbol{z}_i\right) \log \frac{p\left(\boldsymbol{z}_i, \boldsymbol{c}_j \mid \mathbf{\Theta}, \mathbf{X}\right)}{Q\left(\boldsymbol{c}_j \mid \boldsymbol{z}_i\right)}, 
\end{aligned}
\label{obj}
\end{equation}
where $Q\left(\boldsymbol{c}_j \mid \boldsymbol{z}_i\right)$ denotes the distribution of latent variable $\boldsymbol{c}_j$ when $\boldsymbol{z}_i$ is observed. The goal of graph structure learning can be reformulated to maximize the function over $\boldsymbol{z}_i$ when $Q\left(\boldsymbol{c}_j \mid \boldsymbol{z}_i\right)$ is estimated. Since the cluster centroids are latent, we introduce the Expectation–Maximization (EM) algorithm to formulate the optimization process.

In the E-step, $\boldsymbol{z}_i$ is fixed and $Q\left(\boldsymbol{c}_j \mid \boldsymbol{z}_i\right)$ can be estimated by K-Means algorithm over all $\boldsymbol{z}_i$. The distribution is estimated by a hard indicator $\hat{Q}\left(\boldsymbol{c}_k \mid \boldsymbol{z}_i\right)=1$ if POI $l_i$ belongs to $k$-th cluster, and $\hat{Q}\left(\boldsymbol{c}_j \mid \boldsymbol{z}_i\right)=0$ for other centroids $\boldsymbol{c}_j$.

In the M-step, we fix $\hat{Q}\left(\boldsymbol{c}_j \mid \boldsymbol{z}_i\right)$ and optimize $\boldsymbol{z}_i$. By introducing hard indicator $\hat{Q}\left(\boldsymbol{c}_j \mid \boldsymbol{z}_i\right)$, maximizing the lower bound in \cref{obj} yields a loss function:
\begin{equation}
\begin{aligned}
 \mathcal{L}_{\text{HSL}} = - \sum_{l_i \in L} \sum_{\boldsymbol{c}_j \in C} \hat{Q}\left(\boldsymbol{c}_j \mid \boldsymbol{z}_i\right) \log p\left(\boldsymbol{z}_i, \boldsymbol{c}_j \mid \mathbf{\Theta}, \mathbf{X}\right),
\end{aligned}
\end{equation}
Taking inspiration from NCL~\cite{NCL}, we assume that the distribution of POIs is isotropic Gaussian over their corresponding clusters and each Gaussian distribution has the same variance. Therefore, the loss function can be written as:
\begin{equation}
\begin{aligned}
\mathcal{L}_{\text{HSL}}=& -\sum_{l_i \in L} \log \frac{\exp \left(-\left(\boldsymbol{z}_i-\boldsymbol{c}_k\right)^{\top}\cdot \left(\boldsymbol{z}_i-\boldsymbol{c}_k\right) / 2 \sigma^2\right)}{\sum_{\boldsymbol{c}_j \in C} \exp \left(-\left(\boldsymbol{z}_i-\boldsymbol{c}_j\right)^{\top}\cdot\left(\boldsymbol{z}_i-\boldsymbol{c}_j\right) / 2 \sigma^2\right)} \\
\propto&-\sum_{l_i \in L}\log \frac{\exp \left(\boldsymbol{z}_i^{\top} \boldsymbol{c}_k / \tau_1\right)}{\sum_{\boldsymbol{c}_j \in C} \exp \left(\boldsymbol{z}_i^{\top} \boldsymbol{c}_j / \tau_1\right)},
\end{aligned}
\label{eq:temperature1}
\end{equation}
where $\boldsymbol{c}_k$ is the centroid of the cluster to which POI $l_i$ belongs, and $2\sigma^2$ is represented by a temperature coefficient $\tau_1$.
Since $\boldsymbol{z}_i$ and $\boldsymbol{c}_j$ have been $l_2$-normalized in advance, we can leverage $\left(\boldsymbol{z}_i-\boldsymbol{c}_j\right)^{\top}\cdot\left(\boldsymbol{z}_i-\boldsymbol{c}_j\right)= 2-2\boldsymbol{z}_i^{\top} \boldsymbol{c}_j$ to simplify the loss function. 

This objective suggests that in the M-step, the structure embedding of each POI and its corresponding cluster centroid should be as close as possible. We achieve this by iteratively conducting K-Means in the E-step and minimizing $\mathcal{L}_{HSL}$ in the M-step.
With the above objective, the hierarchical nature in POI features can be captured in the structure embedding and the prototypes can be obtained by averaging POI representations in each cluster. The prototype will be used to represent the coarse-grained semantics of clusters and to extend the neighborhoods of POI nodes.

Compared to heuristic grouping approaches, such as grouping temporal features by date, our clustering-based method enables the adaptive uncovering of intricate hierarchical structures in POI features, which may be irregular and not aligned with pre-defined heuristic rules.

After hierarchical structure learning, we can define an adjacency matrix $\mathbf{A}^{\mathrm{Hier}} \in \mathbb{R}^{N \times K}$ between POIs and prototypes based on the hard indicator:
\begin{equation}
    \mathbf{A}^{\mathrm{Hier}}_{i j} = \hat{Q}\left(\boldsymbol{c}_j \mid \boldsymbol{z}_i\right) = \left\{\begin{array}{ll}
1, & \text{$l_i$ belongs to $j$-th cluster}, \\
0, & \text{otherwise}. \\
\end{array}\right.
\end{equation}
\subsection{Pairwise Structure Learning}

Although hierarchical structure information has been explicitly captured, 
the GNNs are still susceptible to the presence of data noise and incompleteness in the graph structures.
Traditional rule-based graph construction relies on pre-defined rules or assumptions, inevitably leading to noisy and missing edges.
In contrast, graph structure learning methods can alleviate these issues by automatically identifying patterns and inferring the topological structure.

To address the issue of noisy and incomplete graphs, we adopt a deep graph structure learning method to learn pairwise relationships between POIs adaptively:
\begin{equation}
\boldsymbol{z}_i = \mathbf{W}_2 \sigma_s \left( \mathbf{W}_1 \boldsymbol{x}_i + \boldsymbol{b}_1\right) + \boldsymbol{b}_2, \label{gsl1}
\end{equation}
\begin{equation}
\mathbf{A}^{\mathrm{POI}}_{i j}=\frac{\boldsymbol{z}_i^{\top} \boldsymbol{z}_j}{\left\|\boldsymbol{z}_i\right\|  \left\|\boldsymbol{z}_j\right\|},
\label{gsl2}
\end{equation}
where $\boldsymbol{x}_i \in \mathbb{R}^{d_1}$ is the primitive feature (e.g., latitude and longitude in the spatial view) of POI $l_i$, which is used to construct graphs, $\boldsymbol{z}_i \in \mathbb{R}^{d_2}$ is the structure embedding transformed by learnable parameters $\mathbf{W}_1 \in \mathbb{R}^{d_2 \times d_1}, \mathbf{W}_2 \in \mathbb{R}^{d_2 \times d_2}, \boldsymbol{b}_1, \boldsymbol{b}_2 \in \mathbb{R}^{d_2}$. $\mathbf{A}^{\mathrm{POI}} \in \mathbb{R}^{N \times N}$ is the adjacency matrix of POI graph. To refine $\mathbf{A}^{\mathrm{POI}}$ into a sparse and normalized adjacency matrix, we also conduct $\epsilon$-neighborhood sparsification and normalization post-processing, which are widely used in graph structure learning~\cite{IDGL,HGSL,SUBLIME}.

Distinct from the previous methods that only consider the resemblance between POIs, we also capture the relationships between coarse-grained prototypes to further explore the hierarchical structure. Following the pairwise structure learning defined in Eq. (\ref{gsl1}) and (\ref{gsl2}), we take the cluster centroids as the primitive features to construct the connections between the prototypes:
\begin{equation}
\tilde{\boldsymbol{c}}_i = \mathbf{W}_4 \sigma_s \left( \mathbf{W}_3 \boldsymbol{c}_i + \boldsymbol{b}_3\right) + \boldsymbol{b}_4, \label{gsl3}
\end{equation}
\begin{equation}
\mathbf{A}^{\mathrm{Proto}}_{i j}=\frac{\tilde{\boldsymbol{c}}_i^{\top} \tilde{\boldsymbol{c}}_j}{\left\|\tilde{\boldsymbol{c}}_i\right\|  \left\|\tilde{\boldsymbol{c}}_j\right\|}.
\label{gsl4}
\end{equation}

Following bi-level graph structure learning, we obtained bi-level graphs that encompass both fine-grained POI-level information and coarse-grained prototype-level information. The prototype-level information unveils the collective characteristics of similar POIs. A conventional graph can be symbolized as $\mathcal{G}=(\mathcal{V}, \mathbf{A})$, where $\mathcal{V}$ represents the set of nodes, and $\mathbf{A}$ denotes the adjacency matrix. 
We present the formalized notation for a bi-level graph as $\mathcal{G}=(\mathcal{V}^{\mathrm{POI}}, \mathcal{V}^{\mathrm{Proto}}, \mathbf{A}^{\mathrm{POI}}, \mathbf{A}^{\mathrm{Proto}}, \mathbf{A}^{\mathrm{Hier}})$, where $\mathcal{V}^{\mathrm{POI}}$ and $\mathcal{V}^{\mathrm{Proto}}$ respectively signify the sets of POI nodes and prototype nodes, and $\mathbf{A}^{\mathrm{POI}} \in \mathbb{R}^{N \times N}, \mathbf{A}^{\mathrm{Proto}} \in \mathbb{R}^{K \times K}, \mathbf{A}^{\mathrm{Hier}} \in \mathbb{R}^{N \times K}$ are the three adjacency matrices learned through bi-level graph structure learning.
These matrices portray the interconnections between POI-POI, prototype-prototype, and POI-prototype, respectively.

\subsection{Multi-Relational Graph Attention Network}
\begin{figure}
    \centering
    \includegraphics[width=0.76\linewidth]{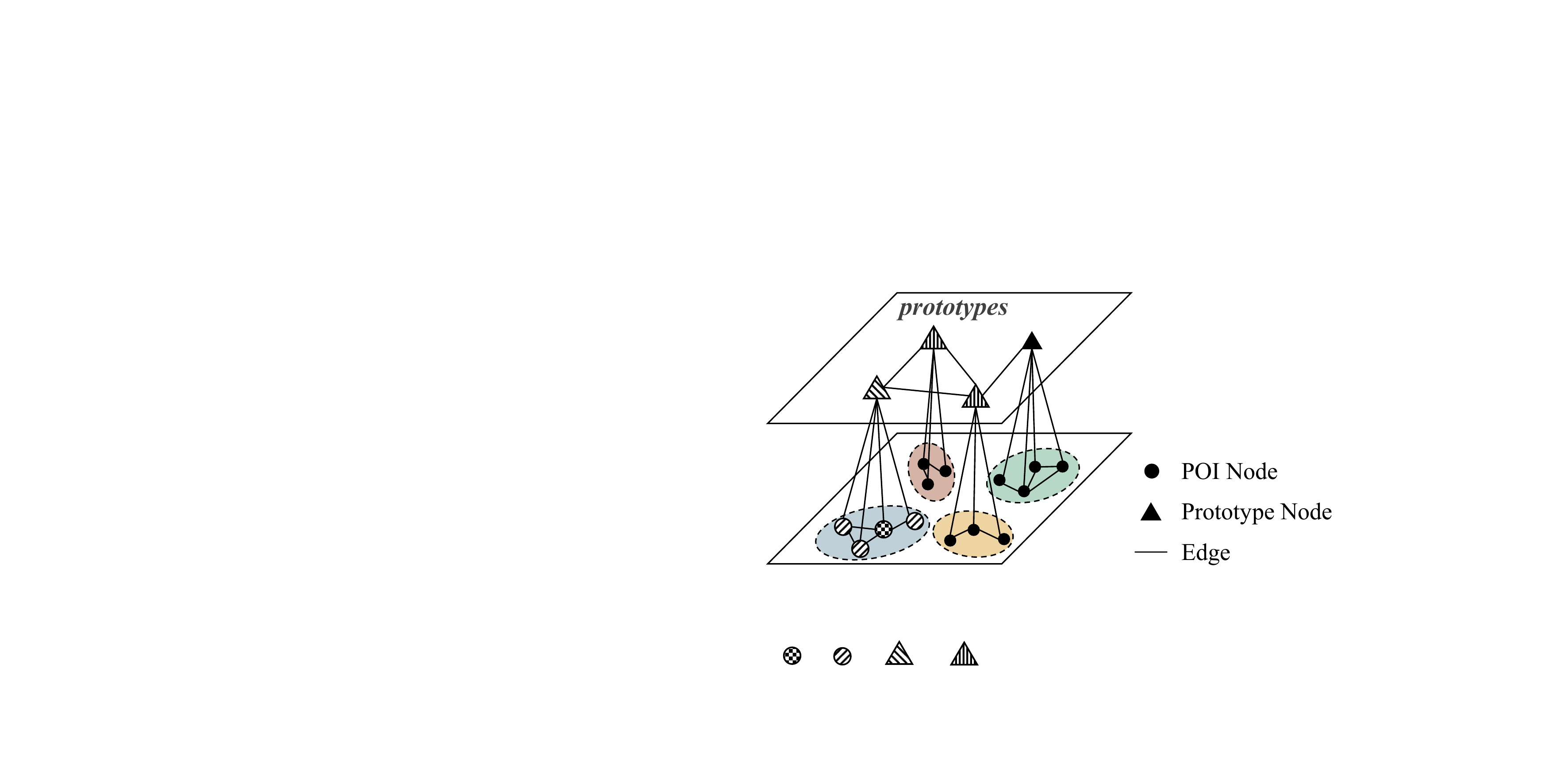}
    \caption{The neighborhood of target node ``\includegraphics[height=2ex]{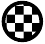}'' for graph representation learning. We define three types of neighbor nodes: ``\includegraphics[height=2ex]{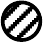}'' denotes the POI-level neighbor nodes, ``\includegraphics[height=2ex]{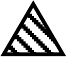}" and ``\includegraphics[height=2ex]{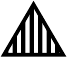}" denote the 1-hop and 2-hop prototype-level neighbor nodes, respectively.}
    \label{neighbor}
\end{figure}

The hierarchical and pairwise structure learning methods produce high-quality hierarchical graphs, which not only are confronted with less noise and missing information, but also manifest the hierarchical information in POI features through the graph topology. 

To fully exploit the learned hierarchical graphs, we design a multi-relational graph attention network.
First, we define the neighborhood of each POI node based on the hierarchical graph. As shown in \cref{neighbor}, we define three types of neighbor nodes. As ``\inlinegraphics{./figures/icons/icon_target_node.pdf}'' is the target node to be updated, ``\inlinegraphics{./figures/icons/icon_poi_neigh.pdf}'' represents the connected POI nodes that are most semantically similar to the target node, ``\inlinegraphics{./figures/icons/icon_1hop_proto_neigh.pdf}" is the prototype of cluster that the target node belongs to, which can provide coarse-grained information about this group of similar nodes, and ``\inlinegraphics{./figures/icons/icon_2hop_proto_neigh.pdf}" represents 2-hop prototype neighbors that provide information about more potential similar groups. 
Thus, there are three types of relations between target node and its neighbor nodes: $\mathcal{R} = \{\inlinegraphics{./figures/icons/icon_target_node.pdf}-\inlinegraphics{./figures/icons/icon_poi_neigh.pdf},\inlinegraphics{./figures/icons/icon_target_node.pdf}-\inlinegraphics{./figures/icons/icon_1hop_proto_neigh.pdf}, \inlinegraphics{./figures/icons/icon_target_node.pdf}-\inlinegraphics{./figures/icons/icon_2hop_proto_neigh.pdf}\}$.
Then, the graph representation learning is defined as:
\begin{equation}
\alpha_{i j}=\frac{\exp \left(\phi\left(\boldsymbol{a}_1^{\top}\left[\mathbf{W}_r \boldsymbol{l}_{i} || \mathbf{W}_r \boldsymbol{l}_j\right]\right)\right)}{\sum_{l_k \in \mathcal{N}^r_i} \exp \left(\phi\left(\boldsymbol{a}_1^{\top}\left[\mathbf{W}_r \boldsymbol{l}_{i} || \mathbf{W}_r \boldsymbol{l}_k\right]\right)\right)},
\label{attention}
\end{equation}
\begin{equation}
\boldsymbol{p}_{i}=\sum_{r\in \mathcal{R}} \sum_{l_j \in \mathcal{N}^r_i}  s_{i j} \alpha_{i j} \mathbf{W}_r \boldsymbol{l}_j + \mathbf{W}_s \boldsymbol{l}_i,
\end{equation}
where $\boldsymbol{l}_{i} \in \mathbb{R}^{d_2}$ denotes the ID embedding of POI $l_{i}$, $\boldsymbol{p}_{i} \in \mathbb{R}^d_3$ is the hidden representation learned from the bi-level graph, and $\mathcal{N}^r_i$ denotes the neighborhood of $l_{i}$ under relation $r\in \mathcal{R}$. $\mathbf{W}_r \in \mathbb{R}^{d_3 \times d_2}$ and $\boldsymbol{a}_1 \in \mathbb{R}^{2d_3}$ are learnable weights, and $\phi(\cdot)$ denotes LeakyReLU activation function. 
$W_r$ is subscripted with $r\in \mathcal{R}$, indicating that aggregation patterns at different levels are modeled separately.
The information propagation from $\boldsymbol{l}_j$ to $\boldsymbol{l}_{i}$ is controlled by attention weight $\alpha_{i j}$ and topology score $s_{i j}$, which is related to learned graph structure:
\renewcommand{\arraystretch}{1.4}
\begin{equation}
s_{i j}=\left\{\begin{array}{ll}
\mathbf{A}^{\mathrm{POI}}_{ij}, & j \text{ is POI neighbor}, \\
\mathbf{A}^{\mathrm{Hier}}_{ij},& j \text{ is 1-hop prototype neighbor}, \\
\mathbf{A}^{\mathrm{Proto}}_{p j}, & j \text{ is 2-hop prototype neighbor}, \\
\end{array}\right.
\label{score}
\end{equation}
where $l_{i}$ belongs to the $p$-th cluster when $j$ is $2$-hop prototype neighbor.
\renewcommand{\arraystretch}{1.0}

\subsection{Contrastive Multiview Fusion}

In POI recommendation, multiple features are widely used to construct graphs, such as spatial graph, temporal graph, and transition graph. We claim that each feature provides a feature view. In each view, a set of POI representations can be obtained by conducting graph representation learning.
POI representations derived from multiple views convey both shared and complementary information.
However, existing methods for fusing representations tend to rely on simple techniques such as summation or concatenation. As a result, important informative features may be overlooked. 

In deep learning, linear disentanglement has been used to obtain distinguishable and generalizable representations~\cite{DG,ModalityRec}.
To fully exploit the information embedded in multiple views, we propose a contrastive multiview fusion method that captures both view-shared and view-specific information. 
Each POI representation is decomposed into view-shared and view-specific parts.
View-shared information denotes the common characteristics of POIs across all views.
Extracting view-shared information aids in constructing robust and generalizable POI representations. In contrast, view-specific information reflects the unique attributes of POIs within a specific view. For instance, a pair of POIs might exhibit strong geographical connections but not exhibit the same peak visitation times. By distinguishing view-specific information, our model can more flexibly adapt to the information presented by different views, offering potential for explaining the decision-making process.

To achieve the aforementioned semantic decoupling, we introduce two optimization objectives to guide the semantics of the learnable representations.

First, we employ contrastive learning to extract view-shared representations.
Henceforth, we employ the symbol $\boldsymbol{p}_i^v$ to represent the latent representation of the $i$-th POI, obtained through the multi-relational graph attention network within the feature view $v$.
Contrastive learning aims to enhance the agreement among diverse views of the same data and has proven effective in multiview and multimodal tasks~\cite{contrastive1,contrastive3}. 
In this work, we propose an adaptive contrastive learning auxiliary task to distill the shared information $\boldsymbol{p}_{c, i} \in \mathbb{R}^{d_3}$ from multiple views by maximizing the agreement between POI representations under different views and the fused representations. 
The resulting view-shared loss can be mathematically noted as:
\begin{equation}
\mathcal{L}_{\mathrm{SH}}=-\frac{1}{|L|} \frac{1}{|V|} \sum_{l_i \in L} \sum_{v \in V} I\left(\boldsymbol{p}_i^v, \boldsymbol{p}_{c, i}\right),
\end{equation}
where $I\left(\cdot, \cdot\right)$ represents the mutual information computed using InfoNCE~\cite{InfoNCE}. Specifically, we define $(\boldsymbol{p}_i^v, \boldsymbol{p}_{c, i})$ as positive samples, while all other POI embeddings within the same view $(\boldsymbol{p}_i^v, \boldsymbol{p}_j^v)$ and other fused multiview embeddings $(\boldsymbol{p}_i^v, \boldsymbol{p}_{c,j})_{(j\neq i)}$ are regarded as negative samples:
\begin{equation}
\begin{aligned}
&I\left(\boldsymbol{p}_i^v, \boldsymbol{p}_{c, i}\right) \\
=&\log \frac{e^{\theta\left(\boldsymbol{p}_i^v, \boldsymbol{p}_{c, i}\right) / \tau_2}}{e^{\theta\left(\boldsymbol{p}_i^v, \boldsymbol{p}_{c, i}\right) / \tau_2}+\sum_{j \neq i}\left(e^{\theta\left(\boldsymbol{p}_i^v, \boldsymbol{p}_{c, j}\right) / \tau_2}+e^{\theta\left(\boldsymbol{p}_i^v, \boldsymbol{p}_j^v\right) / \tau_2}\right)},
\end{aligned}
\label{eq:temperature2}
\end{equation}
where $\tau_2$ is the temperature parameter and $\theta(\cdot, \cdot)$ is the critic function implemented as a cosine similarity function.

Capturing the unique features held by each view is also crucial for a comprehensive understanding of the semantics of POIs.
To this end, we extract view-specific representations to complement the view-shared representations. Specifically, the view-specific representations $\boldsymbol{p}_{s, i}^{v} \in \mathbb{R}^d_3$ of each view $v$ are obtained by subtracting the view-shared representations $\boldsymbol{p}_{c, i}$ from the view representation $\boldsymbol{p}_i^{v}$:
\begin{equation}
    \boldsymbol{p}_{s, i}^{v}=\boldsymbol{p}_i^{v}-\boldsymbol{p}_{c, i}.
\end{equation}
To ensure that view-specific representations do not encode shared information, we employ an orthogonality constraint:
\begin{equation}
\mathcal{L}_{\mathrm{SP}}=\frac{1}{|L|} \sum_{l_i \in L} \sum_{v \in V} \sum_{u \in V/v} \left\|\boldsymbol{p}_{s, i}^{v}{}^{\top} \boldsymbol{p}_{s, i}^{u}\right\|^2 .
\end{equation}

Then, we integrate the view-shared and view-specific representations with an attention module. The importance of each representation for POI $l_i$ is formulated as:
\begin{equation*}
\left[\alpha_{c, i}, \alpha_{s, i}^{v_1}, \cdots, \alpha_{s,i}^{v_{|V|}}\right] =\operatorname{softmax}\left(\boldsymbol{a}_2^{\top}\left[\boldsymbol{p}_{c, i}, \boldsymbol{p}_{s, i}^{v_1}, \cdots, \boldsymbol{p}_{s, i}^{v_{|V|}}\right]\right),
\end{equation*}
where $\boldsymbol{a}_2 \in \mathbb{R}^{d_3}$ is the learnable vector. Then, the final fused multiview representation of POI $l_i$ is formulated as:
\begin{equation}
\tilde{\boldsymbol{p}}_i=\alpha_{c, i} \boldsymbol{p}_{c, i}+\alpha_{s, i}^{v_1} \boldsymbol{p}_{s, i}^{v_1} + \cdots + \alpha_{s, i}^{v_{|V|}} \boldsymbol{p}_{s, i}^{v_{|V|}}.
\label{eq:fusion}
\end{equation}

To allow the recommender to perceive the collaborative signals among POIs in diverse feature views, we introduce fused representations obtained in \cref{eq:fusion} into the backbone by combining POIs' original ID embeddings and fused representations. The details can be found in Appendix A.

\subsection{Optimization}
We adopt the cross entropy (CE) loss to compute the ranking of ground-truth next POI, which encourages the prediction of ground truth next POI to be ranked more highly:
\begin{equation}
\mathcal{L}_{\text {CE}}=-\sum_{i=1}^{|D_{train}|} \log \left(\hat{y}_i\right),
\end{equation} 
where $\hat{y}_i$ is the predicted probability of the ground-truth next POI for the $i$-th training sample, and $|D_{train}|$ is the total number of samples in training set.

The overall objective function can be formulated as:
\begin{equation}
\mathcal{L}=\mathcal{L}_{\text{CE}}+\beta_{\text{HSL}} \mathcal{L}_{\text{HSL}}+\beta_{\mathrm{SH}} \mathcal{L}_{\mathrm{SH}}+\beta_{\mathrm{SP}} \mathcal{L}_{\mathrm{SP}},
\label{loss}
\end{equation}
where $\beta_{\text{HSL}}, \beta_{\mathrm{SH}}, \beta_{\mathrm{SP}}$ are hyper-parameters to control the hierarchical structure learning and contrastive multiview fusion.

Since hierarchical structure learning needs to be performed iteratively, the model is optimized with the EM algorithm. To help better understand the optimization process, we provide the detailed workflow for EM-based optimization in Algorithm~\ref{algo}.

\begin{algorithm}[t]
\caption{EM-based optimization process of \themodel}
\label{algo}
\Input{Training set $D_{train}$, test set $D_{test}$, POI set $L$, feature view set $V$, multiview POI features $\{\boldsymbol{x}^{v}_i\}_{v \in V, l_i\in L}$, and number of clusters $K$.}
\Output{Recommended next POIs for test samples.}
// Training \\
Init POI ID embeddings $\{\boldsymbol{l}_i\}_{l_i\in L}$. \\
Init POI structure embeddings $\{\boldsymbol{z}^v_i\}_{v \in V, l_i\in L}$ with $\{\boldsymbol{x}^{v}_i\}_{v \in V,l_i\in L}$. \\
\While{not converge}{
    \For{minibatch data $d_3$ in $D_{train}$}
    {
        // E-step \\
        \For{feature view $v$ in $V$}
        {
            // Hierarchical Structure Learning
            $\mathbf{A}^{\mathrm{Hier}}_v, \{\boldsymbol{c}^v_j\}_{j=1,\ldots,K}$ = K-Means($\{\boldsymbol{z}^v_i\}_{l_i\in L}$) 
        }
        // M-step \\
        \For{feature view $v$ in $V$}
        {
            // Pairwise Structure Learning \\
            Learn $\mathbf{A}^{\mathrm{POI}}_v, \mathbf{A}^{\mathrm{Proto}}_v$ via \cref{gsl1} to (\ref{gsl4}). \\
            // Multi-Relational Graph Learning \\
            Learn $\{\boldsymbol{p}^v_i\}_{l_i\in L}$ via \cref{attention} to (\ref{score}). \\
        }
        // Contrastive Multiview Fusion \\
        Fuse multiview representations $\{\boldsymbol{p}^v_i\}_{v\in V,l_i\in L}$ to $\{\tilde{\boldsymbol{p}}_i\}_{l_i\in L}$ via \cref{eq:fusion}. \\
        Predict next POIs $\hat{\boldsymbol{y}}$ with the recommender. \\
        Calculate loss $\mathcal{L}$ via \cref{loss}. \\
        Update parameters by applying gradient descent. \\
    }
}
// Testing \\
Predict next POIs for test samples in $D_{test}$ with the recommender and learned graph representations. \\
Return predicted next POIs for test samples.
\end{algorithm}

\subsection{Complexity Analysis}
Our design encompasses hierarchical structure learning based on the EM algorithm, pairwise structure learning based on metric learning,  multi-relational GNN, and contrastive multiview fusion. For our proposed bi-level graph, we use \(N\) and \(K\) to represent the number of POIs and prototypes, respectively. \(E_1\), \(E_2\), and \(E_3\) are used to denote the number of edges in the adjacency matrices \(\mathbf{A}^{\mathrm{POI}}\), \(\mathbf{A}^{\mathrm{Hier}}\), and \(\mathbf{A}^{\mathrm{Proto}}\), respectively.
We analyze the complexity of each component individually:
\begin{enumerate}
    \item Hierarchical Structure Learning: We employ the EM algorithm to learn the hierarchical structure within POI features. In the E-step, we execute K-Means to assign each POI to its corresponding group, and in the M-step, we update the structural representation of the POIs by optimizing \(\mathcal{L}_{\text{HSL}}\). In both the E-step and M-step, we need to compute the pairwise distances between POIs and cluster centroids, resulting in a complexity of \(O(KN + KN) = O(KN)\).
    \item Pairwise Structure Learning: We learn the POI-level pairwise structure by computing the pairwise similarity between POI nodes. The same operation is also performed on prototype nodes. Therefore, the complexity of this part is \(O(N^2 + K^2)\).
    \item Multi-Relational GNN: We modify the vanilla GNN to accommodate our bi-level graph, which includes \(N+K\) nodes and \(E_1+E_2+E_3\) edges. 
    Since the complexity of a vanilla GNN layer is $O(N+E)$, where $N$ and $E$ are the number of nodes and edges respectively,
    we can conclude that the complexity of our modified graph model is \(O(N+E_1+E_2+E_3)\). Since we only update the representations of POI nodes, the complexity is independent of \(K\).
    \item Contrastive Multiview Fusion: To implement the representation decomposition and fusion we proposed, we need to calculate \(\mathcal{L}_{\mathrm{SH}}\) and \(\mathcal{L}_{\mathrm{SP}}\). The complexity of computing these two losses is \(O(VN^2)\) and \(O(V^2N)\), respectively. Since the number of views \(V\) is a very small constant, the complexity of this part is \(O(VN^2 + V^2N) = O(N^2)\).
\end{enumerate}
Based on the above analysis, we can conclude that the total complexity of our design is \(O(KN + N^2 + K^2 + N + E_1 + E_2 + E_3)\).
In our model, the following inequalities hold:
\begin{itemize}
    \item \(K \ll N\) (the number of prototypes is much less than the number of POIs)
    \item \(N < E_1\) (the degree of each POI node is greater than 1)
    \item \(E_1 < N^2\) (POIs are not fully connected)
    \item \(E_2 = N\) (each POI belongs to only one cluster)
    \item \(E_3 < E_1\) (the number of prototype-prototype edges is less than the number of POI-POI edges)
\end{itemize}
Thus, the total complexity of our design can be simplified as $O(N^2)$.
The efficiency bottlenecks lie in the POI-level pairwise structure learning and the contrastive fusion, which involves the scalability issues of graph structure learning and contrastive learning.
In fact, several methods have been proposed to enhance the efficiency of these two techniques~\cite{NodeFormer,GGD}. 
These methods provide valuable support for further enhancing efficiency.

%% file: sections/experiments.tex
\section{Experiments}

In this section, we first introduce the evaluation setups and then present the empirical results to enable a fair comparison. Subsequently, we provide detailed analyses of ablation, sensitivity, actual runtime, exploration ability, and visualizations to validate the effectiveness of our \themodel.

\subsection{Experimental Setups}

\subsubsection{Datasets}

We evaluate our proposed \themodel model on three real-world datasets: Gowalla, Foursquare, and BJ. 
\textbf{Gowalla}\footnote{\url{http://snap.stanford.edu/data/loc-gowalla.html}}~\cite{Gowalla} contains check-ins of users over the period of February 2009 - October 2010 from all over the world.
\textbf{Foursquare}\footnote{\url{https://sites.google.com/site/yangdingqi/home/foursquare-dataset}}~\cite{Foursquare} contains check-ins in 415 cities collected from April 2012 to September 2013.
\textbf{BJ} dataset is provided by Meituan, which includes real-world transaction records in Beijing. The dataset is sampled from the logs of a mobile application for the period spanning from 1 October 2021 to 25 October 2021. We regard the user's order payment behavior as the check-in in traditional location-based social networks.

We follow the protocol of HMT-GRN~\cite{HMT-GRN} to keep users with check-in counts between 20 and 50, and remove POIs that have been visited by fewer than 10 users. The numbers of users, POIs, and check-ins in each dataset and data density after preprocessing are shown in \cref{dataset}. 
After sorting the timestamps in chronological order,  we use the first 80\% visits and the last 20\% visits of each user’s sequence for training and testing respectively.

\begin{table}[t]
    \caption{Dataset Statistics\label{dataset}}
	\centering
	\begin{tabular}{crrrc}
		\toprule
    Dataset& \multicolumn{1}{c}{\#Users} & \multicolumn{1}{c}{\#POIs} & \multicolumn{1}{c}{\#Check-ins} & Density \\
    \midrule
    Gowalla    & 11,864& 3,359 & 86,670 & 0.168\%\\
    Foursquare & 16,636& 4,455 & 170,573 & 0.155\%\\
    BJ         & 6,096 & 6,032& 148,736 & 0.275\%\\
    \bottomrule
	\end{tabular}
\end{table}

\begin{table*}[t]
  \caption{Performance comparison with baselines. The best performance is highlighted in \textbf{bold} and the runner-up is highlighted by \underline{underlines}. Improvement indicates relative improvements over the best baseline in percentage.}
  \label{performance}
  \centering
    \resizebox{\linewidth}{!}{
    \begin{tabular}{ccccccccccccccc} 
      \toprule
      & \multicolumn{4}{c}{Gowalla} & &  \multicolumn{4}{c}{Foursquare} & & \multicolumn{4}{c}{BJ} \\
      \cmidrule(lr){2-5}  \cmidrule(lr){7-10} \cmidrule(lr){12-15}
      & Acc@5 & Acc@10 & Acc@20 & MRR & & Acc@5 & Acc@10 & Acc@20 & MRR & & Acc@5 & Acc@10 & Acc@20 & MRR \\ 
      \midrule
         RNN             & 0.1873  & 0.2440  & 0.3050 & 0.1381 &  & 0.2246  & 0.2973  & 0.3752 & 0.1700 &  &  0.1553  & 0.2182  & 0.2839 & 0.1076 \\ 
         GRU             & 0.1869  & 0.2489  & 0.3161 & 0.1406 &  & 0.2300  & 0.3027  & 0.3852 & 0.1740 &  &  0.1570  & 0.2224  & 0.2929 & 0.1098 \\ 
         LSTM            & 0.1968  & 0.2575  & 0.3276 & 0.1510 &  & 0.2437  & 0.3174  & 0.4032 & 0.1854 &  &  0.1658  & 0.2316  & 0.3050 & 0.1175 \\ 
         HST-LSTM        & 0.0366  & 0.0636  & 0.1004 & 0.0279 &  & 0.0307  & 0.0500  & 0.0806 & 0.0244 &  &  0.0250  & 0.0447  & 0.0741 & 0.0181 \\ 
         STGCN           & 0.0909  & 0.1351  & 0.1955 & 0.0684 &  & 0.0948  & 0.1531  & 0.2323 & 0.0703 &  &  0.0701  & 0.1165  & 0.1789 & 0.0486 \\ 
         LSTPM           & 0.2282  & 0.2720  & 0.3200 & 0.1803 &  & 0.2671  & 0.3214  & 0.3778 & 0.2078 &  &  0.1870  & 0.2396  & 0.2919 & 0.1360 \\ 
         STAN            & 0.1928  & 0.2440  & 0.3039 & 0.1460 &  & 0.2382  & 0.3136  & 0.3987 & 0.1759 &  &  0.1623  & 0.2240  & 0.2918 & 0.1126 \\ 
         STP-UDGAT       & 0.2374  & 0.2783  & 0.3202 & 0.1770 &  & 0.2926  & 0.3556  & 0.4187 & 0.2136 &  &  0.1996  & 0.2548  & 0.0538 & 0.1366 \\ 
         Flashback       & 0.2342  & 0.2770  & 0.3285 & 0.1821 &  & 0.2768  & 0.3347  & 0.4012 & 0.2118 &  &  0.1928  & 0.2467  & 0.3047 & 0.1380 \\ 
         GETNext         & 0.2546  & 0.3008  & 0.3683 & 0.1950 &  & 0.3141  & 0.3806  & 0.4566 & 0.2352 &  &  0.2142  & 0.2778  & 0.3390 & 0.1505 \\ 
         Graph-Flashback & 0.2593  & 0.3040  & 0.3730 & 0.1979 &  & 0.3272  & 0.3994  & 0.4607 & 0.2399 &  &  0.2205  & 0.2820  & 0.3479 & 0.1530 \\ 
         SNPM            & 0.2658  & 0.3234  & 0.3952 & 0.2091 &  & 0.3304  & 0.4135  & 0.4978 & 0.2493 &  &  0.2231  & 0.2889  & 0.3507 & 0.1542 \\ 
         HMT-GRN & \underline{0.2783}  & \underline{0.3394}  & \underline{0.4033} & \underline{0.2120}  &  & \underline{0.3357}  & \underline{0.4148}  & \underline{0.4983} & \underline{0.2510} &  & \underline{0.2285} & \underline{0.3010} & \underline{0.3606} & \underline{0.1564}  \\ 
      \midrule
         \rowcolor[gray]{.8} \themodel & \textbf{0.2923}  & \textbf{0.3685}  & \textbf{0.4471} & \textbf{0.2162}  &  & \textbf{0.3500}  & \textbf{0.4423}  & \textbf{0.5323} & \textbf{0.2552}  &  & \textbf{0.2422} & \textbf{0.3271} & \textbf{0.4095} & \textbf{0.1650}  \\ 
         Improvement & 5.04\% & 8.57\% & 10.85\% & 1.96\% &  & 4.25\% & 6.62\% & 6.81\% & 1.67\%  &  & 5.94\% & 8.68\% & 13.56\%  & 5.48\% \\ 
      \bottomrule
    \end{tabular}
}
\end{table*}

\subsubsection{Baselines}

To show the effectiveness of our proposed model, we compare it with the following baseline models:

\begin{itemize}
\item \textbf{RNN} is a recurrent network that captures sequential dependencies in check-in sequences but suffers from the vanishing gradient issue. The variants of \textbf{GRU} and \textbf{LSTM} introduce gate mechanism to control information flow and alleviate the above issue.
\item \textbf{HST-LSTM}~\cite{HST-LSTM} incorporates spatial and temporal intervals between check-ins into the LSTM gates. Similarly, \textbf{STGCN}~\cite{STGCN} models the intervals with new gates.
\item \textbf{LSTPM}~\cite{LSTPM} is an LSTM-based model that effectively captures both short-term and long-term user preferences by employing a geo-dilated operation for the former and a geo-nonlocal operation for the latter.
\item \textbf{STAN}~\cite{STAN} introduces dependencies of non-adjacent POIs and non-consecutive visits into the self-attention model, to model the dependencies more effectively.
\item \textbf{STP-UDGAT}~\cite{STP-UDGAT} is a graph-based model that captures global correlations among POIs and personalized user preferences through multiple graphs.
\item \textbf{Flashback}~\cite{Flashback} accesses historical hidden states with similar contexts, thus utilizing rich spatio-temporal information from sparse user mobility traces.
\item \textbf{GETNext}~\cite{GETNext} introduces a global trajectory graph to leverage the extensive collaborative signals globally.
\item \textbf{Graph-Flashback}~\cite{Graph-Flashback} incorporates the POI graph learned from spatial-temporal knowledge graph into RNNs for capturing the transition patterns.
\item \textbf{SNPM}~\cite{SNPM} adopts the Eigenmap method to construct a latent POI similarity graph to tackle the sparsity issue.
\item \textbf{HMT-GRN}~\cite{HMT-GRN} is a graph-based model that constructs global spatio-temporal POI graphs to model collaborative signals. It also utilizes auxiliary next-region prediction tasks to alleviate the data sparsity issue.
\end{itemize}

\subsubsection{Metrics}

The performance is evaluated by how well the target POIs in the candidate set are ranked. We adopt two widely-used metrics of ranking evaluation: $\text{Acc@}K$, which counts the fraction of times that the target POI is among the top $K$ probability samples, and Mean Reciprocal Rank (MRR).
The details of metrics can be found in Appendix B.

\subsubsection{Settings}

Adam is employed as the optimizer, where the learning rate is set to 1e-4. The training process would be stopped after 60 epochs. The batch size is set to 96. The dimension of POI/user ID embedding $d_2$ and hidden dimension $d_3$ are both set to $1024$. The number of clusters $K=80$ for Gowalla and $K=120$ for Foursquare and BJ. The temperature coefficient $\tau_1$ is $0.1$ and $\tau_2$ is $0.5$. The weight of loss function $\beta_{\mathrm{HSL}},\beta_{\mathrm{C}},\beta_{\mathrm{S}}$ are set to 1e-4, 1e-1, 1e-4 respectively. 
We use geolocation as the primitive spatial feature and time slot frequency distribution as the primitive temporal feature (a week is partitioned equally into 56 time slots).

\subsection{Performance Comparison}

We compare \themodel with the baselines and the results are summarized in \cref{performance}. Under all the metrics, \themodel can significantly outperform all the baselines on each of the datasets, which demonstrates its effectiveness in improving next POI recommendation.

Among baseline models, graph-based models, such as GETNext, Graph-Flashback, SNPM, and HMT-GRN, perform better than other models. It is not a surprise since graph-based models are better at capturing global POI-POI relationships and transition patterns across users. 

GETNext, Graph-Flashback, SNPM, and HMT-GRM are the best-performing baseline models, they still have some deficiencies compared to our proposed \themodel. HMT-GRN takes into account the hierarchical structure in spatial features by introducing objectives of auxiliary tasks at the regional levels. However, these objectives in multi-task learning paradigm are independent, therefore fall short in modeling the interaction between different levels.
Additionally, it utilizes spatial and temporal graphs constructed according to pre-defined rules, resulting in performance affected by the noise in the graphs. 
The advantage of our method over GETNext lies in our consideration of graph-based encoding for multiple features, which is more capable of capturing the intricate relationships between POIs than the direct embedding layer used in GETNext.
Graph-Flashback considers learning a latent POI graph with the knowledge graph containing spatio-temporal information, but it does not explicitly consider hierarchical structure when learning latent graphs. 
Although SNPM adopts the Eigenmap method to infer the latent graph structure, it remains constrained by pre-defined rules, which results in inferior performance compared to ours.
Moreover, for fusing multiple features, they either simply perform a linear combination of multiple features or directly embed them into a unified latent graph, leading to suboptimal results and lacking interpretability. For modeling hierarchical structures, even though some baseline models employ stacked hierarchical encoders to extract subsequence-level information from sequences, they are limited to capturing hierarchical patterns in sequence features and are incapable of modeling more comprehensive topological structure between POIs. Due to these limitations in their modeling capabilities, the performance of these methods is not as effective as our proposed method.

Our \themodel achieves the best performance, which unequivocally outperforms all baseline models and verifies the effectiveness of our proposed methods. Compared with existing graph-based models, the \themodel model automatically learns hierarchical structures and pairwise structures embedded in the POI features, and fuses the information mined from them in a more effective manner.

\begin{table}[t]
  \caption{Ablation analysis by removing components.}
  \label{ablation}
  \centering
  \resizebox{\linewidth}{!}{
    \begin{tabular}{cccccc}
      \toprule
      & &  Acc@5 &  Acc@10 & Acc@20 & MRR \\ 
      \midrule
      \multirow{6}{*}{Gowalla} & \cellcolor[gray]{.8} \themodel  & \cellcolor[gray]{.8} \textbf{0.2923} & \cellcolor[gray]{.8} \textbf{0.3685} & \cellcolor[gray]{.8} \textbf{0.4471} & \cellcolor[gray]{.8} \textbf{0.2162}  \\
      ~ & w/o HSL        & 0.2898 & 0.3587 & 0.4293 & 0.2152 \\
      ~ & w/o PSL        & 0.2803 & 0.3435 & 0.4135 & 0.2133 \\
      ~ & w/o Shar     & 0.2904 & 0.3608 & 0.4327 & 0.2158 \\
      ~ & w/o Spec   & 0.2856 & 0.3536 & 0.4274 & 0.2147 \\
      ~ & w/o Shar\&Spec & 0.2837 & 0.3523 & 0.4265 & 0.2143 \\
      \midrule
      \multirow{6}{*}{Foursquare} & \cellcolor[gray]{.8} \themodel  & \cellcolor[gray]{.8} \textbf{0.3500} & \cellcolor[gray]{.8} \textbf{0.4423} & \cellcolor[gray]{.8} \textbf{0.5323} & \cellcolor[gray]{.8} \textbf{0.2552} \\
      ~ & w/o HSL     & 0.3482 & 0.4345 & 0.5179 & 0.2548  \\
      ~ & w/o PSL     & 0.3405 & 0.4220 & 0.5049 & 0.2537  \\
      ~ & w/o Shar     & 0.3488 & 0.4363 & 0.5206 & 0.2550  \\
      ~ & w/o Spec    & 0.3450 & 0.4307 & 0.5164 & 0.2545  \\
      ~ & w/o Shar\&Spec & 0.3434 & 0.4289 & 0.5160 & 0.2543  \\
      \midrule
      \multirow{6}{*}{BJ} & \cellcolor[gray]{.8} \themodel  & \cellcolor[gray]{.8} \textbf{0.2422} & \cellcolor[gray]{.8} \textbf{0.3271} & \cellcolor[gray]{.8} \textbf{0.4095} & \cellcolor[gray]{.8} \textbf{0.1650} \\
      ~ & w/o HSL        & 0.2403 & 0.3173 & 0.3908 & 0.1645  \\
      ~ & w/o PSL        & 0.2310 & 0.3021 & 0.3744 & 0.1607 \\
      ~ & w/o Shar     & 0.2407 & 0.3196 & 0.3945 & 0.1647  \\
      ~ & w/o Spec   & 0.2361 & 0.3120 & 0.3888 & 0.1625  \\
      ~ & w/o Shar\&Spec & 0.2343 & 0.3108 & 0.3878 & 0.1616  \\
      \bottomrule
    \end{tabular}
}
\end{table}

\begin{figure}
\centering
\subfigure[Learned spatial structure embedding (w/o $\mathcal{L}_{\text{HSL}}$)] {
\includegraphics[width=1.32in]{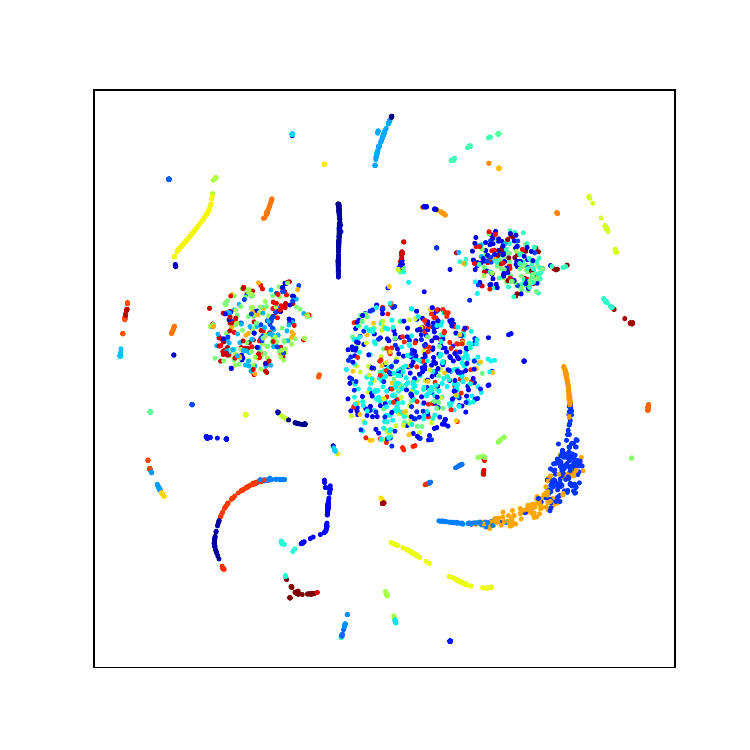}  
}
\subfigure[Learned spatial structure embedding (w/ $\mathcal{L}_{\text{HSL}}$)] {
\includegraphics[width=1.32in]{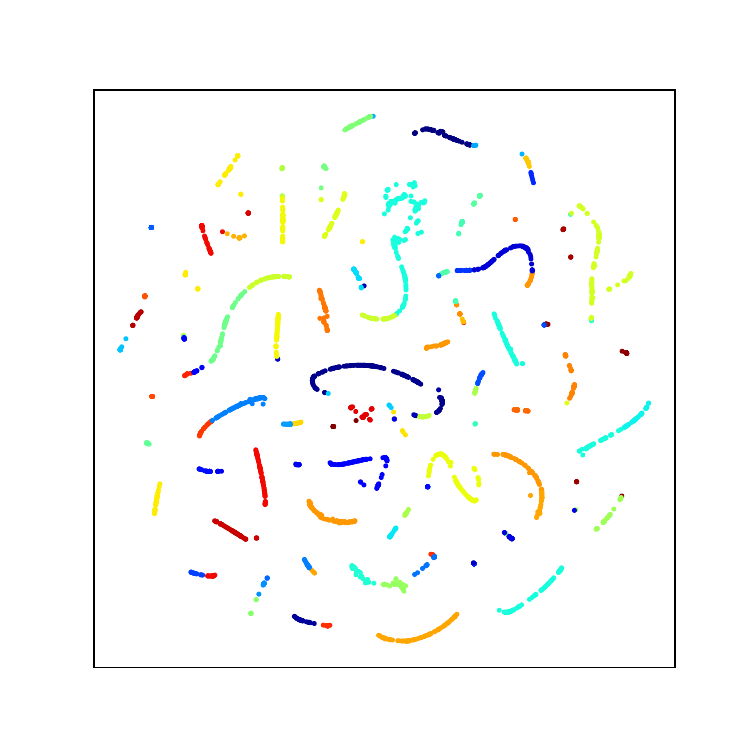}  
}
\subfigure[Learned temporal structure embedding (w/o $\mathcal{L}_{\text{HSL}}$)] {
\includegraphics[width=1.32in]{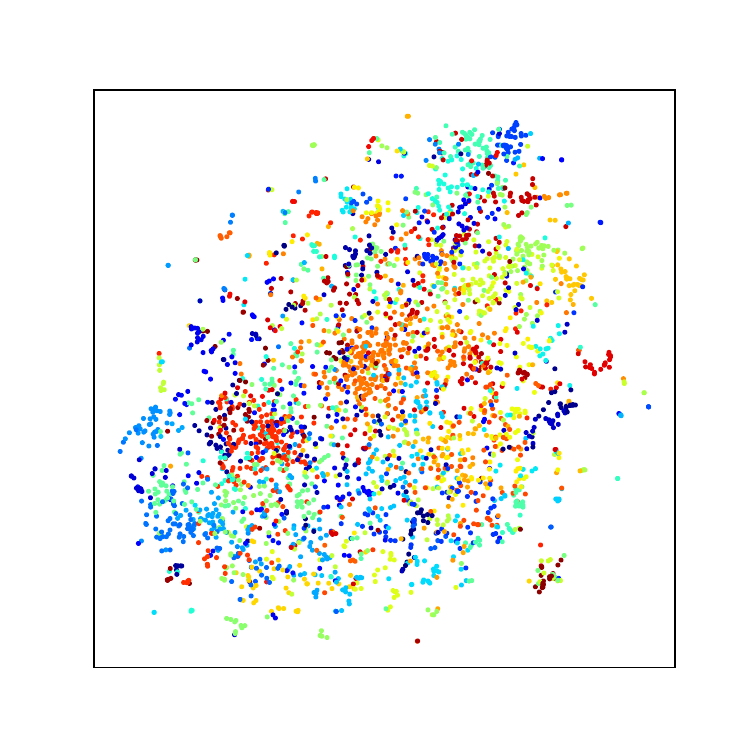}  
}
\subfigure[Learned temporal structure embedding (w/ $\mathcal{L}_{\text{HSL}}$)] {
\includegraphics[width=1.32in]{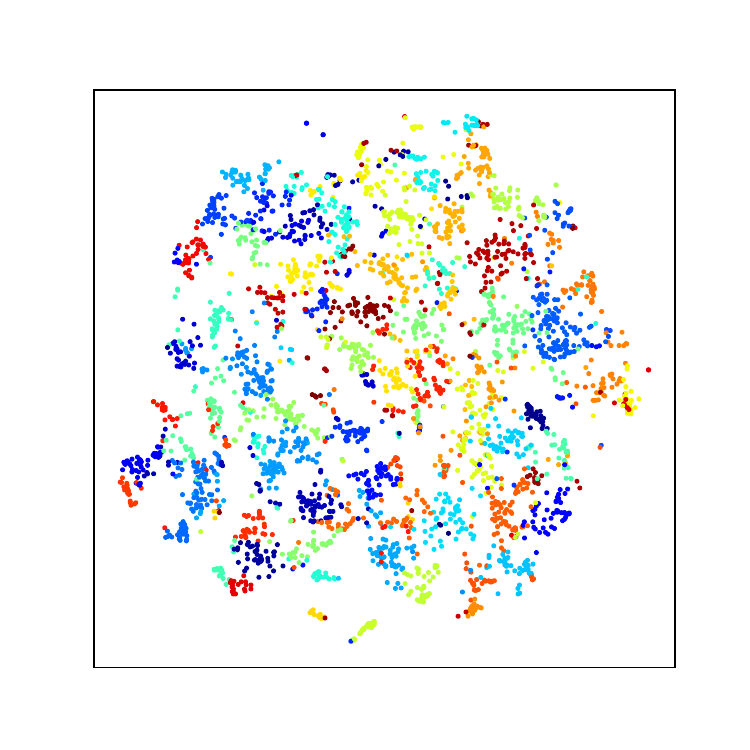}  
}
\caption{The t-SNE visualization of learned structure embedding on the Gowalla dataset. Colors indicate clustering labels. The introduction of $\mathcal{L}_{\text{HSL}}$ can help the model to effectively capture the clustering and hierarchical information in POI features, and the clustering property of spatial features is more pronounced than that of temporal features.}
\label{vis-tsne}
\end{figure}

\subsection{Ablation Study}

To analyze the effectiveness of the different components, we conduct an ablation study. We denote the base model as \themodel and drop different components to form variants. The main components in our model are listed as:

\begin{itemize}
    \item \textbf{\textit{HSL}}: The hierarchical structure learning module, which is responsible for capturing clustering and hierarchical information in POI features and extracting prototypes. Removing this component will result in no more prototype neighbors in graph representation learning.
    \item \textbf{\textit{PSL}}: The pairwise structure learning module, which learns relative relationships between POIs or prototypes adaptively. Removing this module means that we use pre-defined rules for graph construction and the hierarchical structure learning based on the learned structure embedding will not be employed.
    \item \textbf{\textit{Shared (Shar)}}: The view-shared loss $\mathcal{L}_{\mathrm{SH}}$ in the contrastive multiview fusion, which aims to extract information shared by representations in multiple views.
    \item \textbf{\textit{Specific (Spec)}}: The view-specific loss $\mathcal{L}_{\mathrm{SP}}$ in the contrastive multiview fusion, which ensures the view-specific representations of different feature views do not encode view-shared information.
\end{itemize}

The results are summarized in \cref{ablation}. We have the following observations from this table.

We observe that the hierarchical structure learning helps to improve the model performance, verifying the effectiveness of this component. Since this component mainly leverages the loss function $\mathcal{L}_{\mathrm{HSL}}$ to control the optimization process in the EM algorithm, to further understand the influence of this component, we visualize the effect of $\mathcal{L}_{\mathrm{HSL}}$ on graph structure learning in \cref{vis-tsne}. 

From \cref{vis-tsne}, we can intuitively find that the structure embedding learned based on either spatial or temporal features has more obvious clustering under the guidance of $\mathcal{L}_{\mathrm{HSL}}$. This indicates that the introduction of $\mathcal{L}_{\mathrm{HSL}}$ can effectively capture the hierarchical properties in POI features and guide the learning of structure embedding. In addition, the spatial structure embedding has more pronounced clustering than temporal structure embedding, indicating that spatial features naturally exhibit more significant hierarchical structures. Consequently, in graphs constructed with these embeddings, connected POIs are more semantically similar. The prototypes extracted from clusters present more coarse-grained group features, thus constructing effective hierarchical structures.

The results in \cref{ablation} also demonstrate the effectiveness of pairwise structure learning and contrastive multiview fusion. Note that view-specific information has a more significant impact on contrastive fusion than view-shared information, which indicates that the difference in semantics of different features is more pronounced than similarity, and it is meaningful to fully consider multiple features in POI recommendation.

\begin{figure}
  \centering
\subfigure[Varied $K$ on Gowalla]
{
\includegraphics[width=0.44\linewidth]{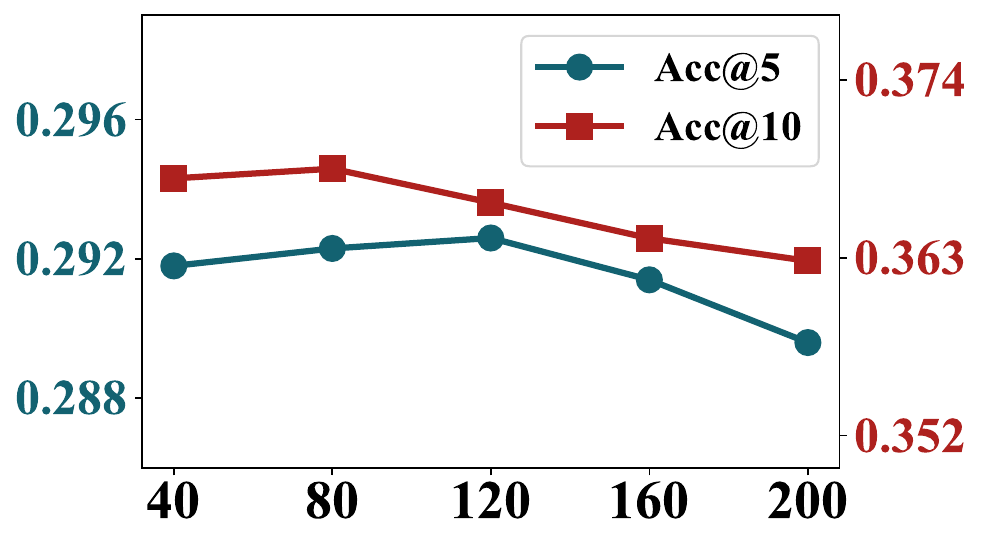}
}
\subfigure[Varied $K$ on Foursquare]
{
\includegraphics[width=0.44\linewidth]{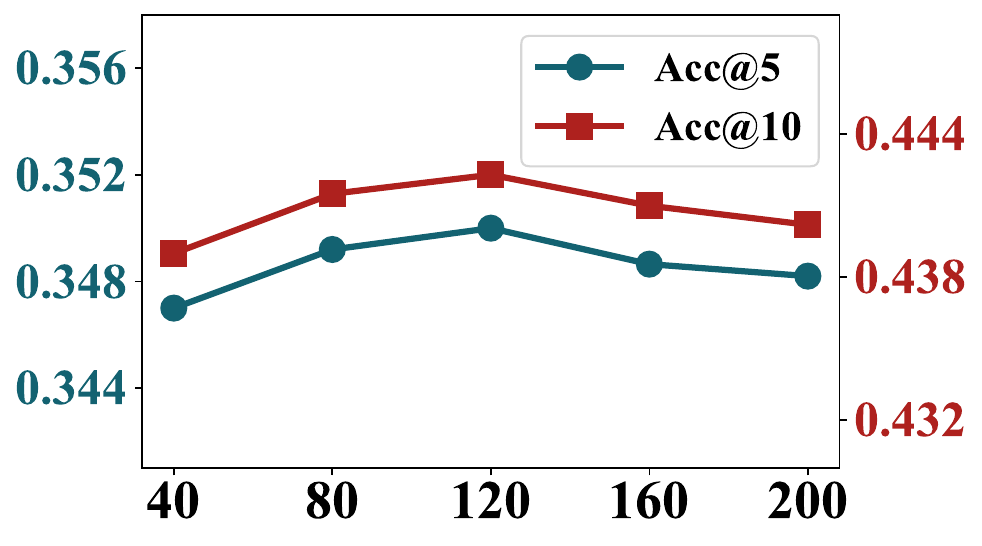}  
}
\subfigure[Varied $\beta_{\text{HSL}}$ on Gowalla]
{
\includegraphics[width=0.44\linewidth]{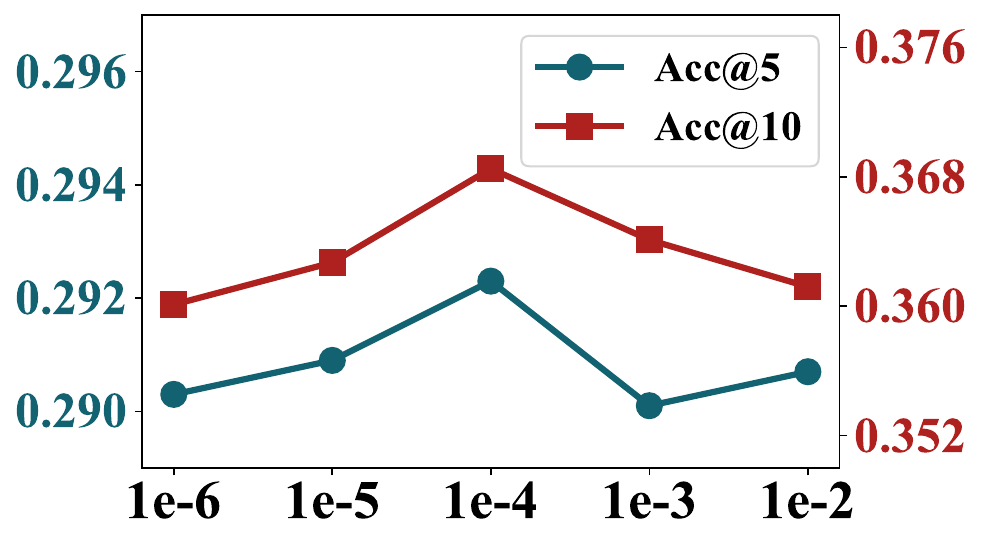}  
}
\subfigure[Varied $\beta_{\text{HSL}}$ on Foursquare]
{
\includegraphics[width=0.44\linewidth]{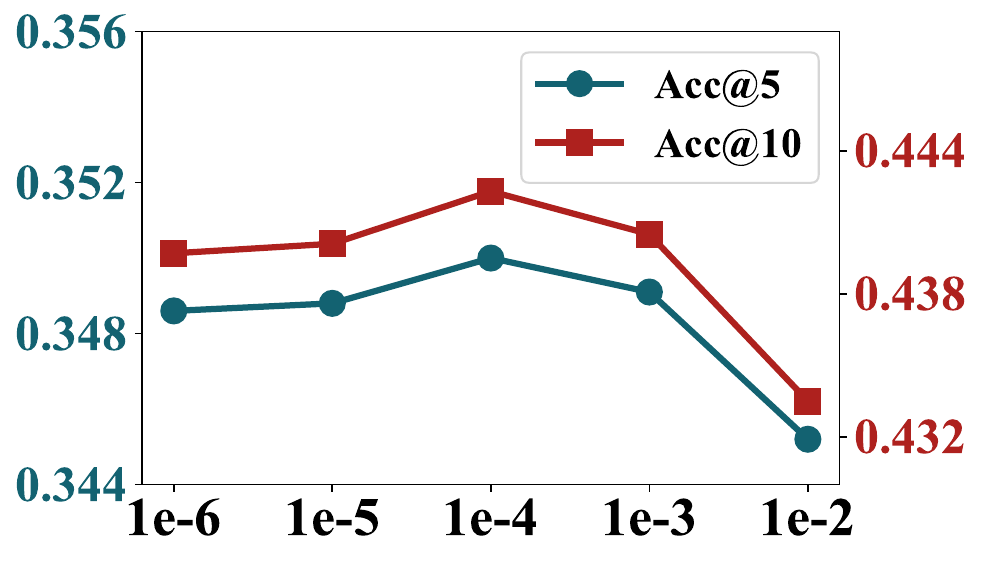}  
}
\subfigure[Varied $\beta_{\text{SH}}$ on Gowalla]
{
\includegraphics[width=0.44\linewidth]{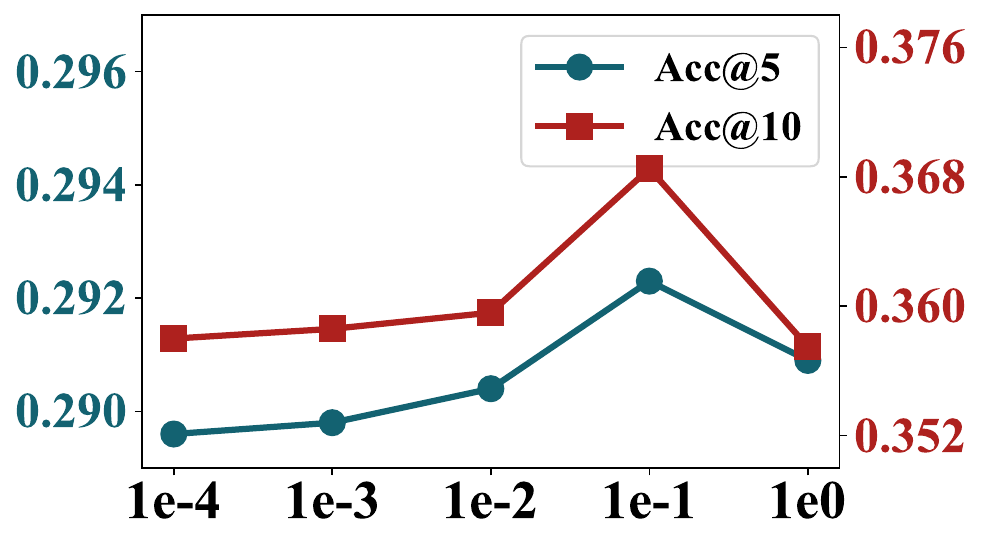}  
}
\subfigure[Varied $\beta_{\text{SH}}$ on Foursquare]
{
\includegraphics[width=0.44\linewidth]{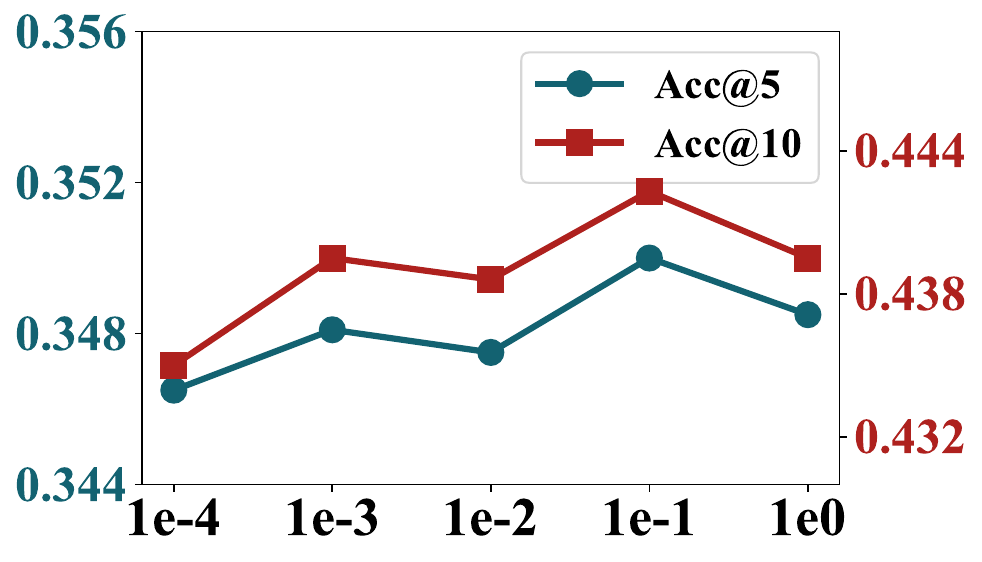}  
}
\subfigure[Varied $\beta_{\text{SP}}$ on Gowalla]
{
\includegraphics[width=0.44\linewidth]{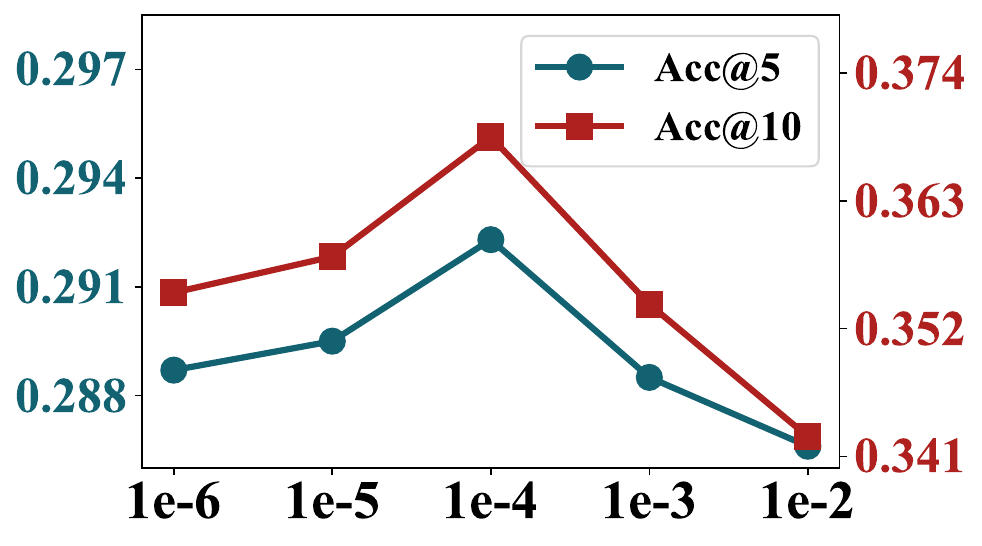}  
}
\subfigure[Varied $\beta_{\text{SP}}$ on Foursquare]
{
\includegraphics[width=0.44\linewidth]{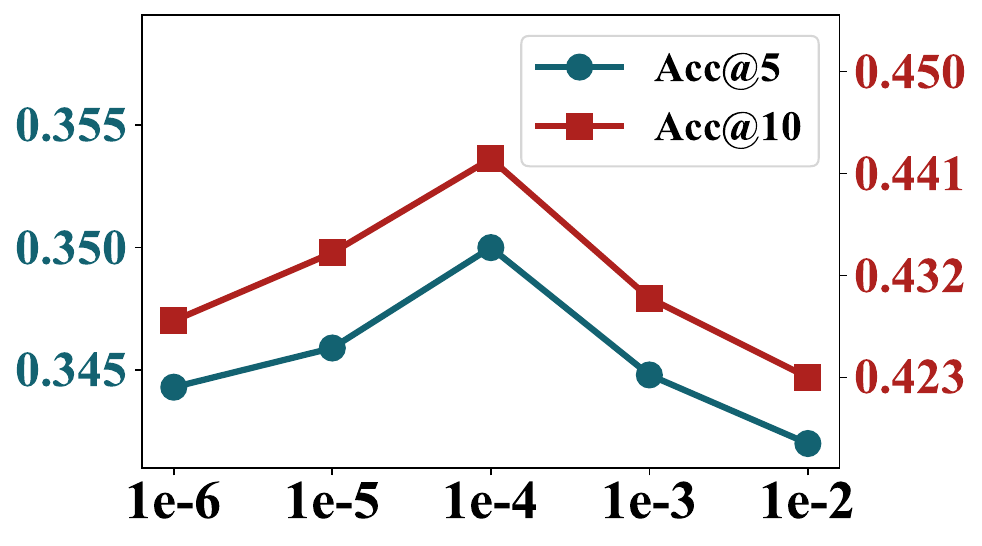}  
}
\caption{Effect of different cluster numbers and loss weights. The y-axis represents accuracy and the x-axis is the different hyper-parameter values.}
\label{sensitivity}
\end{figure}

\subsection{Sensitivity Analysis}

We conduct sensitivity analysis with four hyper-parameters on the hierarchical structure learning and the contrastive multiview fusion, which are the most pivotal parts of \themodel. The investigated hyper-parameters include the number of clusters $K$, the weights of loss function $\beta_{\text{HSL}}$, $\beta_{\text{SH}}$, $\beta_{\text{SP}}$, and the temperature coefficients $\tau_1, \tau_2$. \cref{sensitivity} and \cref{fig:sensitivity-temperature} shows the effect of varied hyper-parameter values, from which we have the following observations.

\subsubsection{Effect of cluster number $K$.}
When excavating hierarchical structure and extracting prototypes, we employ K-Means to conduct node clustering. We vary the cluster number $K$ from 40 to 200 with step 40.
\cref{ablation} (a,b) shows that $K=80$ and $K=120$ are the best cluster numbers on Gowalla and Foursquare datasets respectively. We speculate that the best number of clusters is related to the number of POIs in the dataset. When the number of POIs increases, there may also be more feature patterns in the dataset and more clusters are needed to distinguish them. Since the number of POIs in Foursquare is more than that in Gowalla, Foursquare needs more clusters and prototypes to represent the coarse-grained groups.

\subsubsection{Effect of weights of loss functions $\beta_{\text{HSL}},\beta_{\text{SH}},\beta_{\text{SP}}$.}
We analyze the effect of $\beta_{\text{HSL}}$, $\beta_{\text{SH}}$, and $\beta_{\text{SP}}$ since they determine the effectiveness of hierarchical structure learning and contrastive multiview fusion.
As shown in \cref{ablation}, we tune them to analyze their effect. On both datasets, the best performance is achieved when $\beta_{\text{HSL}}$=1e-4, $\beta_{\text{SH}}$=1e-1, and $\beta_{\text{SP}}$=1e-4. Note that, consistent with the observations of the ablation study, view-specific information has a greater impact on performance than view-shared information.

\begin{figure}[t]
  \centering
\subfigure[Varied $\tau_1$ on Gowalla]
{
\includegraphics[width=0.44\linewidth]{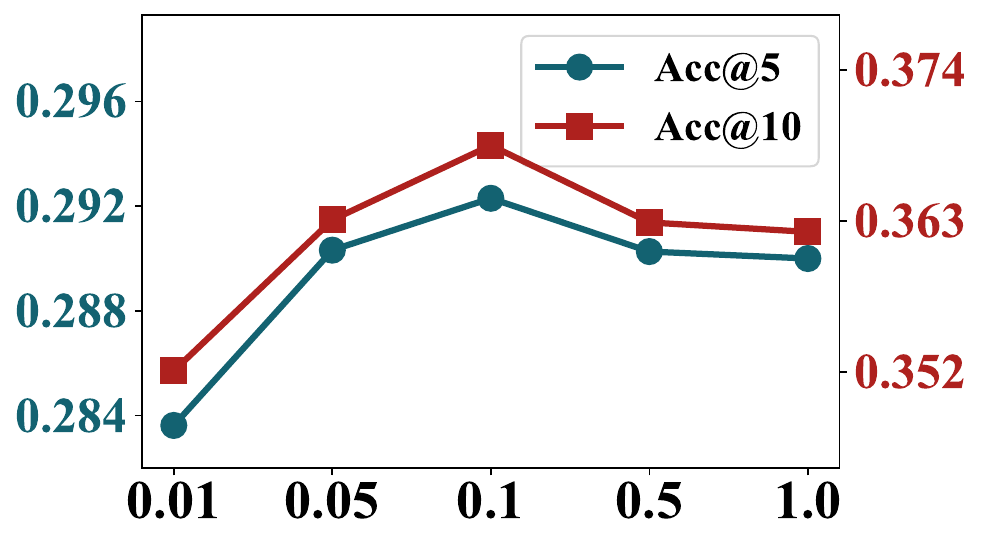}
}
\subfigure[Varied $\tau_1$ on Foursquare]
{
\includegraphics[width=0.44\linewidth]{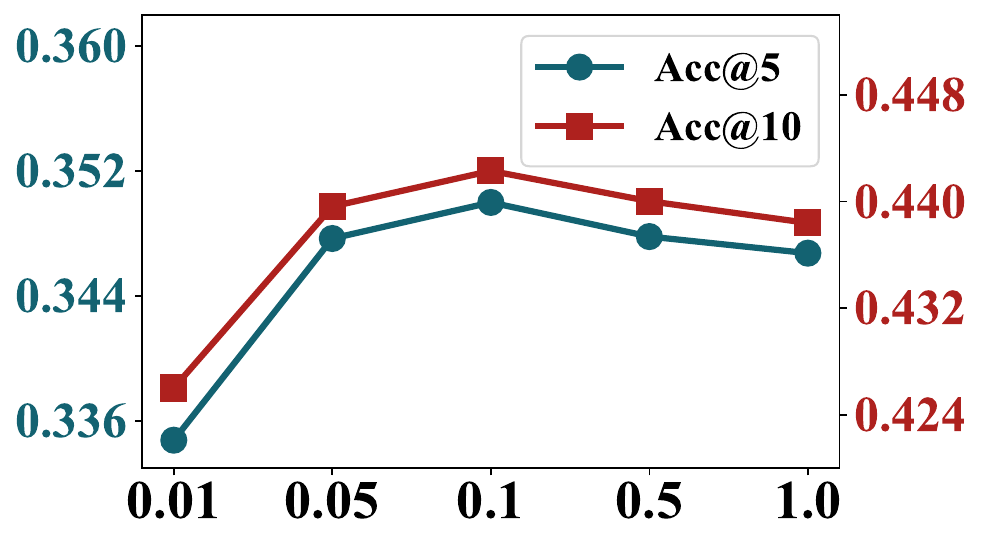}  
}
\subfigure[Varied $\tau_2$ on Gowalla]
{
\includegraphics[width=0.44\linewidth]{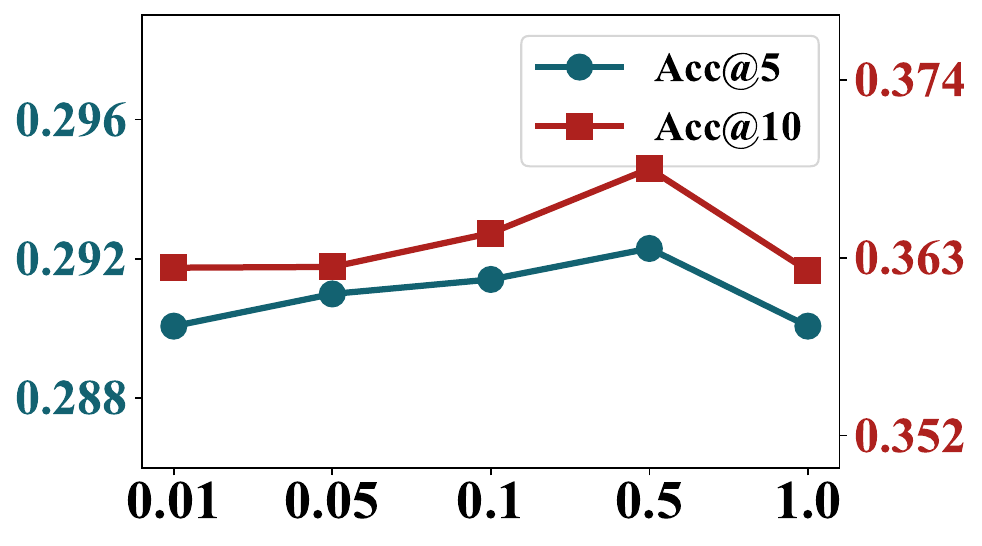}  
}
\subfigure[Varied $\tau_2$ on Foursquare]
{
\includegraphics[width=0.44\linewidth]{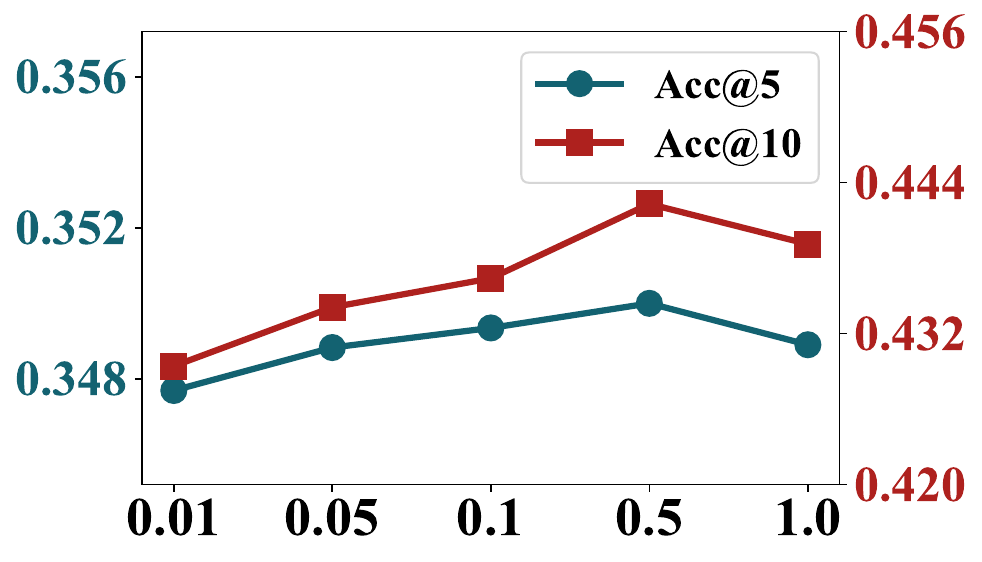}  
}
\caption{Effect of different values of temperature coefficients $\tau_1$ and $\tau_2$. The y-axis represents accuracy and the x-axis is the different temperature coefficient values.}
\label{fig:sensitivity-temperature}
\end{figure}

\subsubsection{Effect of temperature coefficients $\tau_1, \tau_2$.}
Our method incorporates two temperature coefficients: $\tau_1$ in \cref{eq:temperature1}, which controls hierarchical clustering, and $\tau_2$ in \cref{eq:temperature2}, used for computing the mutual information in the view-shared loss. We test the performance of the model with each of them set to different values within [0.01, 0.05, 0.1, 0.5, 1].
The experimental results, as shown in \cref{fig:sensitivity-temperature}, indicate that $\tau_1=0.1$ and $\tau_2=0.5$ are the optimal choices.
From \cref{fig:sensitivity-temperature}, small values of \(\tau_1\) and \(\tau_2\) lead to a significant decrease in performance. This is due to \(\tau_1\) and \(\tau_2\) controlling the smoothness of the similarity computation. Lower temperature coefficients result in reduced smoothness, which enlarges the differences in similarity between different sample pairs. This makes the similarity more sensitive to changes in representations, easily leading to overfitting, thus impairing the model's generalization ability.

\begin{table*}[t]
  \caption{Performance comparison regarding feature grouping scheme. Our method is compared against method without feature grouping and two traditional grouping methods: group spatial features by grid, and group temporal features by date. Our performance is best and is highlighted in \textbf{bold}, and the runner-up is highlighted by \underline{underlines}.}
  \label{tab: feature-grouping}
  \centering
    \resizebox{\linewidth}{!}{
    \begin{tabular}{cccccccccccc} 
      \toprule
      \multirow{2}*{Feature Grouping Scheme} & \multirow{2}*{Group Method} & \multicolumn{4}{c}{Gowalla} & &  \multicolumn{4}{c}{Foursquare} \\
      \cmidrule(lr){3-6}  \cmidrule(lr){8-11}
      ~ & ~ & Acc@5 & Acc@10 & Acc@20 & MRR & & Acc@5 & Acc@10 & Acc@20 & MRR \\ 
      \midrule
        Without Feature Grouping & None & {0.2898} & {0.3587} & {0.4293} & {0.2152} & ~ & \underline{0.3482} & \underline{0.4345} & \underline{0.5179} & \underline{0.2548} \\ 
      \midrule
       \multirow{2}*{\makecell{Pre-defined Rule Based}} & Group Spatial Features by Grid & {0.2826} & {0.3563} & {0.4276} & {0.2072} & ~ & {0.3359} & {0.4274} & {0.5138} & {0.2477} \\ 
       ~ & Group Temporal Features by Date & \underline{0.2901} & \underline{0.3611} & \underline{0.4308} & \underline{0.2154} & ~ & {0.3476} & {0.4294} & {0.5157} & {0.2502} \\ 
      \midrule
        \makecell{Ours (Clustering Based)} & Hierarchical Structure Learning & \textbf{0.2923}  & \textbf{0.3685}  & \textbf{0.4471} & \textbf{0.2162}  &  & \textbf{0.3500}  & \textbf{0.4423}  & \textbf{0.5323} & \textbf{0.2552} \\ 
      \bottomrule
    \end{tabular}
}
\end{table*}

\begin{table*}[t]
  \caption{Performance comparison regarding fusion strategies. Our method is compared against three groups of representative fusion strategies: early, intermediate, and late fusion strategies. Our performance is best and is highlighted in \textbf{bold}, while the best performance in each group of compared strategies is highlighted by \underline{underlines}.}
  \label{tab: feature-fusion}
  \centering
    \resizebox{\linewidth}{!}{
    \begin{tabular}{cccccccccccc} 
      \toprule
      \multirow{2}*{\makecell{Representation Fusion\\ Scheme}} & \multirow{2}*{Fusion Method} & \multicolumn{4}{c}{Gowalla} & &  \multicolumn{4}{c}{Foursquare} \\
      \cmidrule(lr){3-6}  \cmidrule(lr){8-11}
      ~ & ~ & Acc@5 & Acc@10 & Acc@20 & MRR & & Acc@5 & Acc@10 & Acc@20 & MRR \\ 
      \midrule
       \multirow{4}*{Early Fusion} & Concatenate & 0.2794 & 0.3550 & 0.4345 & 0.2024 & ~ & 0.3344 & 0.4173 & 0.5051 & 0.2470 \\ 
       ~ & Linear Sum ($\lambda_1=0.3$) & 0.2841 & 0.3620 & 0.4370 & 0.2096 & ~ & 0.3388 & 0.4279 & 0.5194 & 0.2479 \\ 
       ~ & Linear Sum ($\lambda_1=0.5$) & 0.2856 & 0.3598 & 0.4380 & 0.2096 & ~ & \underline{0.3417} & \underline{0.4323} & \underline{0.5248} & \underline{0.2491} \\
       ~ & Linear Sum ($\lambda_1=0.7$) & \underline{0.2890} & \underline{0.3642} & \underline{0.4423} & \underline{0.2111} & ~ & 0.3400 & 0.4311 & 0.5216 & 0.2489 \\
       \midrule
       \multirow{4}*{Intermediate Fusion} & Concatenate & 0.2702 & 0.3416 & 0.4242 & 0.2015 & ~ & 0.3236 & 0.4109 & 0.4998 & 0.2339 \\
       ~ & Linear Sum ($\lambda_2=0.3$) & 0.2845 & 0.3620 & 0.4400 & 0.2105 & ~ & 0.3383 & 0.4299 & 0.5225 & 0.2452 \\
       ~ & Linear Sum ($\lambda_2=0.5$) & 0.2852 & 0.3640 & 0.4426 & 0.2106 & ~ & \underline{0.3409} & \underline{0.4319} & \underline{0.5234} & \underline{0.2494} \\
       ~ & Linear Sum ($\lambda_2=0.7$) & \underline{0.2870} & \underline{0.3645} & \underline{0.4464} & \underline{0.2108} & ~ & 0.3399 & 0.4307 & 0.5235 & 0.2463 \\
       \midrule
       {Late Fusion} & Average & \underline{0.2807} & \underline{0.3529} & \underline{0.4276} & \underline{0.2072} & ~ & \underline{0.3343} & \underline{0.4221} & \underline{0.5140} & \underline{0.2450} \\ 
      \midrule
        Ours & \makecell{Contrastive Decomposition\\ + Attentive Fusion} & \textbf{0.2923}  & \textbf{0.3685}  & \textbf{0.4471} & \textbf{0.2162}  &  & \textbf{0.3500}  & \textbf{0.4423}  & \textbf{0.5323} & \textbf{0.2552} \\ 
      \bottomrule
    \end{tabular}
}
\end{table*}

\subsection{Comparison of Feature Grouping Schemes}

We adopt a clustering-based grouping method in \cref{sec:hierarchical} to construct hierarchical relationships.
Regarding the feature grouping scheme, we compare our clustering-based grouping with the approach of not conducting feature grouping (i.e., not modeling the hierarchical structure). We also compare changing the grouping of spatial features from clustering-based to grid-based grouping, and the grouping of temporal features from clustering-based to date-based grouping. Experiments are conducted on Gowalla and Foursquare, with the results presented in \cref{tab: feature-grouping}.

From \cref{tab: feature-grouping}, we observe that compared to not conducting feature grouping, traditional grouping schemes do not always improve performance. An improvement is only noted on Gowalla when temporal features are grouped by date, but this still falls short of our clustering-based grouping method. This indicates that the hierarchical structure within POI features is complex and nonlinear, making it challenging to be directly modeled by heuristic rules.
Conversely, our method is capable of adaptively modeling the complex hierarchical structure within POI features through the K-Means clustering and EM algorithm framework.

\subsection{Comparison of Representation Fusion Schemes}

For the representation fusion scheme~\cite{multimodal-survey,fusion2}, we compared our contrastive fusion with other strategies such as early, intermediate, and late fusion for integrating spatial and temporal representations. In early fusion, we perform a concatenation or a linear sum on the node representations obtained in each view. In the linear sum, the weights for spatial and temporal representations are \(\lambda_1\) and \(1-\lambda_1\), respectively. Similarly, in intermediate fusion, we perform a concatenation or a linear sum on the intermediate results after further encoding with the LSTM layer. In the linear sum, the weights for spatial and temporal representation are set as \(\lambda_2\) and \(1-\lambda_2\). In late fusion, we average the predicted logits from different views at the decision layer. Comparative experiments are conducted on Gowalla and Foursquare, with the results presented in \cref{tab: feature-fusion}.

From \cref{tab: feature-fusion}, we observe that among the compared fusion schemes, early fusion performs the best, while late fusion yields the worst results. This may be due to the high correlation between POI representations from different views, where early fusion facilitates better interaction between representations. Additionally, we notice that linear sum outperforms concatenation, possibly because concatenation increases the hidden layer dimension, leading to over-parameterization due to the introduction of more model parameters. Lastly, when evaluating the performance of linear sum, the optimal value of \(\lambda_1\) or \(\lambda_2\) varies across different datasets. This indicates that the optimal weights for linear sum relies on empirical manual selection, whereas our method could adaptively learn the optimal weights.

\subsection{Actual Runtime Analysis}
We compare the actual runtime of our method with the best baseline, HMT-GRN, by recording the total training time required for convergence.
All experiments are conducted on a NVIDIA RTX 3090 GPU to ensure fairness in comparison.
The runtime required to train HMT-GRN and our method is around 5.2 hours and 6.5 hours, respectively.
We observe that due to the difference in the complexity of the training process, our method requires more time per batch compared to HMT-GRN.
However, our design does not lead to a significant increase in actual runtime due to the efficient implementation and the PyTorch library's automatic acceleration of matrix multiplication operations.
Therefore, in terms of total training time, our method is comparable to existing methods.

\subsection{Exploration Ability Analysis}

\begin{table}[t]
  \caption{Performance of next \textit{new} POI recommendation on the BJ dataset.}
  \label{exploration}
    \centering
    \resizebox{\linewidth}{!}{
    \begin{tabular}{cccccc} 
      \toprule
      &  $N^2$-Acc@5 &  $N^2$-Acc@10 & $N^2$-Acc@20 & $N^2$-MRR \\ 
      \midrule
      HMT-GRN & 0.0809 & 0.1190 & 0.1739 & 0.0617 \\
      \midrule
     \themodel  & \textbf{0.1002} & \textbf{0.1380} & \textbf{0.1881} & \textbf{0.0721}  \\
      Improvement  & 23.88\% & 15.87\% & 8.11\% & 16.67\% \\
      \bottomrule
    \end{tabular}
}
\end{table}

In recommendation, balancing personalization and exploration is a critical challenge. Personalization focuses on learning preferences from POIs which users have already visited. In contrast, exploration evaluates the ability to accurately recommend POIs that users have not visited. Both abilities ensure accuracy and diversity in recommendation.

To evaluate the exploration performance of POI recommendation, we adopt the \textit{Next New} ($N^2$) extension of metrics (i.e., $N^2$-Acc@K, $N^2$-MRR) proposed in HMT-GRN~\cite{HMT-GRN}. This set of metrics only considers \textit{Next New} POIs that have never been visited by the user, and thus can be used to evaluate the exploratory ability of the recommenders.

The higher-density dataset may provide more global collaborative signals for exploration, so we conduct experiments on the BJ dataset. The results are shown in the \cref{exploration}. HMT-GRN is currently the most state-of-the-art model for next new POI recommendation, thanks to its selectivity layer designed to identify whether a POI has been visited by a user.
Compared to HMT-GRN, our model significantly enhances the exploration ability without hurting the personalization performance. Note that we do not employ the selectivity layer proposed by HMT-GRN, but only rely on the graph structure learning and the multi-relational graph network to achieve it. The improvement of the $N^2$ metric score illustrates that our method can effectively establish connections for POIs with similar feature semantics and explore more potential POIs in recommendation.

\begin{figure}
\centering
\subfigure[Pre-defined rules]
{
\includegraphics[width=1.5in]{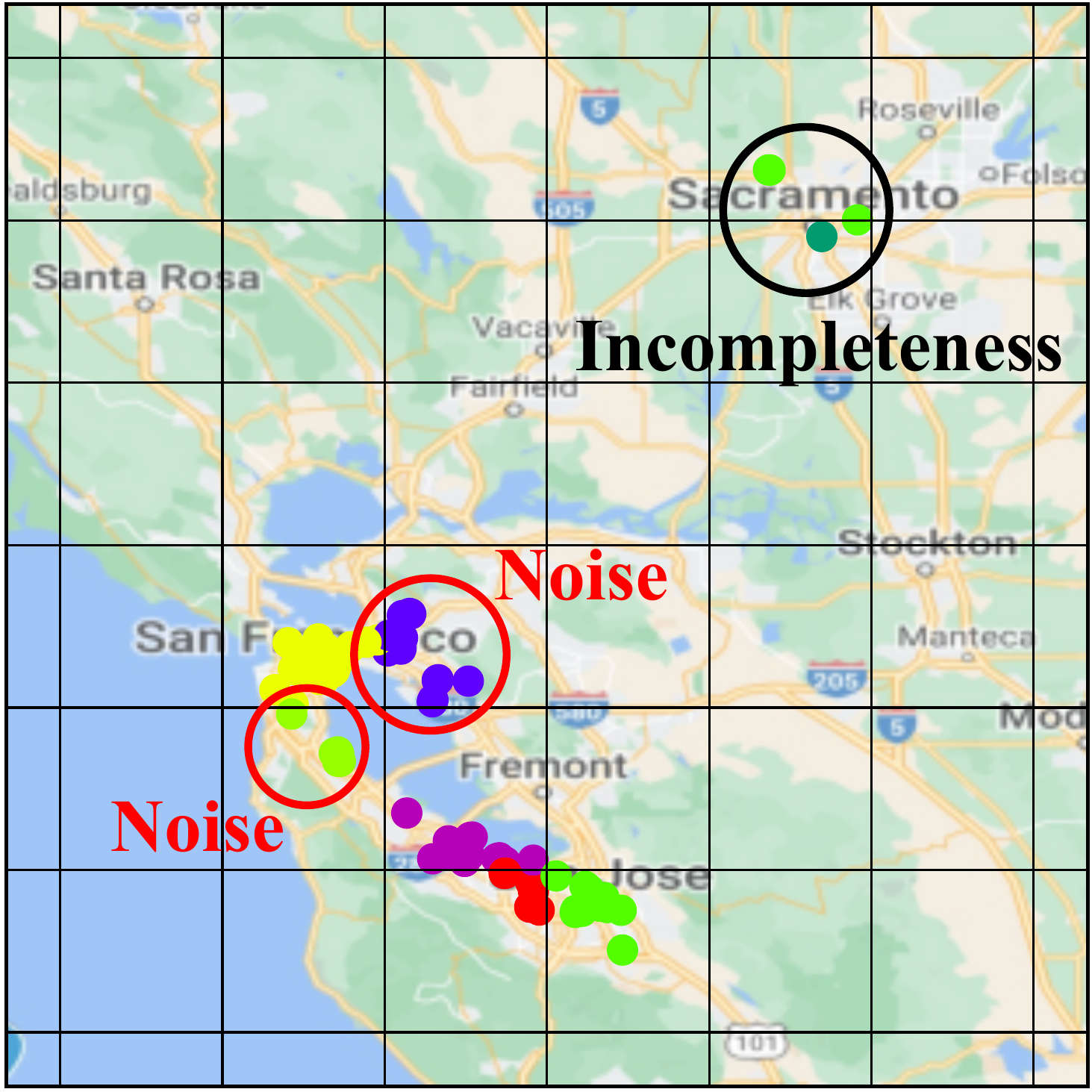}  
}
\subfigure[Ours]
{
\includegraphics[width=1.5in]{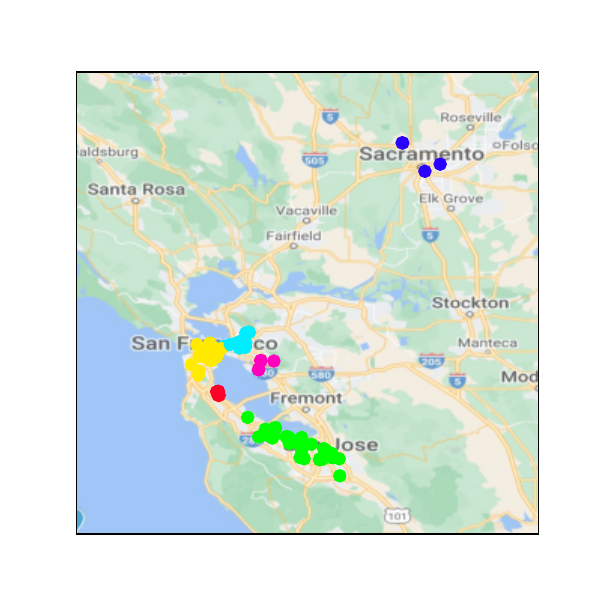}  
}
\caption{Visualization of spatial graphs constructed by pre-defined rules and our method. POIs with the same color are connected in graphs. The graph constructed by pre-defined rules is prone to noise (in the red circles) and incompleteness (in the black circle), while our method alleviates these issues.}
\label{vis}
\end{figure}

\subsection{Visualization of Graph Structure Learning}

In \cref{vis}, we visualize an example of the spatial graphs obtained with pre-defined rules (e.g. grid mapping) and our structure learning method, respectively. The POIs with the same color in the figure are connected by edges. For brevity, we only show the colors of the POI nodes here instead of plotting the edges. Intuitively, the graph constructed by pre-defined rules is prone to noise (in the red circles) and incompleteness (in the black circle), i.e., geographically non-aggregated POIs may be connected by edges, while geographically aggregated POIs are grouped into different grids. In contrast, our method adaptively learns a better spatial graph with less uncertainty. The higher graph quality is helpful for the learning of POI representations and downstream POI recommendations.

%% file: sections/conclusion.tex
\section{Conclusion}

In this study, we propose a novel bi-level graph structure learning method for next POI recommendation, named \themodel.
We employ hierarchical and pairwise structure learning to automatically learn hierarchical graphs, addressing the lack of hierarchical information and the challenge posed by noisy connections.
Based on the learned hierarchical graphs, we devise a novel multi-relational graph attention network to capture collaborative relationships among POIs, considering both POI-level and prototype-level neighbors.
Furthermore, we propose a contrastive multiview fusion strategy to facilitate information fusion.
Comprehensive comparison with baseline models unequivocally shows the superior performance of \themodel.